\documentclass[pdflatex,sn-mathphys-num,iicol]{sn-jnl}

\usepackage{graphicx}%
\usepackage{xspace}
\usepackage{amsmath,amssymb,amsbsy,amsfonts,dsfont,pifont,bm,bbm,mathrsfs,mathtools,nicefrac}
\usepackage{amsthm}%
\usepackage{mathrsfs}%
\usepackage[title]{appendix}%
\usepackage{xcolor}%
\usepackage{textcomp}%
\usepackage{manyfoot}%
\usepackage{algorithm}%
\usepackage{float,wrapfig}
\usepackage{booktabs,multirow,adjustbox,diagbox,threeparttable,tabularray,makecell,caption}
\usepackage{algorithm,algpseudocode,listings}
\usepackage{enumitem}

\theoremstyle{thmstyleone}%
\theoremstyle{thmstyletwo}%

\theoremstyle{thmstylethree}%

\newcommand{\newtext}[1]{{\color{black}#1}}

\begin{document}

\title[Article Title]{DeltaSpace: A Semantic-aligned Feature Space for Flexible Text-guided Image Editing}


\author[1]{\fnm{Yueming} \sur{Lyu}}\email{ymlv@nju.edu.cn}

\author[2]{\fnm{Kang} \sur{Zhao}}\email{zhankang.zk@alibaba-inc.com}

\author[3]{\fnm{Bo} \sur{Peng}}\email{bo.peng@nlpr.ia.ac.cn}

\author[1]{\fnm{Huafeng} \sur{Chen}}\email{nju.chf.111@gmail.com}

\author[3]{\fnm{Yue} \sur{Jiang}}\email{jiangyue2021@ia.ac.cn}

\author[2]{\fnm{Yingya} \sur{Zhang}}\email{yingya.zyy@alibaba-inc.com}

\author*[3]{\fnm{Jing} \sur{Dong}}\email{jdong@nju.edu.cn}

\author*[1]{\fnm{Caifeng} \sur{Shan}}\email{cfshan@nju.edu.cn}

\author[1,3]{\fnm{Tieniu} \sur{Tan}}\email{tnt@nju.edu.cn}

\affil[1]{\orgdiv{PRLab}, \orgname{Nanjing University}, \orgaddress{\city{Suzhou}, \postcode{215163}, \country{China}}}

\affil[2]{\orgdiv{Tongyi Lab}, \orgname{Alibaba Group}, \orgaddress{\city{Beijing}, \postcode{100102}, \country{China}}}

\affil[3]{\orgdiv{Institute of Automation}, \orgname{Chinese Academy of Sciences}, \orgaddress{\city{Beijing}, \postcode{100190}, \country{China}}}


\abstract{
Text-guided image editing faces significant challenges when considering training and inference flexibility. Much literature collects large amounts of annotated image-text pairs to train text-conditioned generative models from scratch, which is expensive and not efficient. After that, some approaches that leverage pre-trained vision-language models have been proposed to avoid data collection, but they are limited by either per text-prompt optimization or inference-time hyper-parameters tuning. To address these issues, we investigate and identify a specific space, referred to as CLIP DeltaSpace, where the CLIP visual feature difference of two images is semantically aligned with the CLIP textual feature difference of their corresponding text descriptions. Based on DeltaSpace, we propose a novel framework called DeltaEdit, which maps the CLIP visual feature differences to the latent space directions of a generative model during the training phase, and predicts the latent space directions from the CLIP textual feature differences during the inference phase. And this design endows DeltaEdit with two advantages: (1) text-free training; (2) generalization to various text prompts for zero-shot inference. Extensive experiments validate the effectiveness and versatility of DeltaEdit with different generative models, including both the GAN model and the diffusion model, in achieving flexible text-guided image editing. Code is available at \url{https://github.com/Yueming6568/DeltaEdit}.
}




\maketitle

\section{Introduction}\label{sec:introduction}

Text-guided image editing has generated widespread research interests in both academic and industrial communities given its significance for real-world applications. The goal of text-guided image editing is to modify the content of images according to user-provided natural language descriptions while keeping the text-irrelevant content unchanged.

\begin{figure*}[t]
    \centering
    \includegraphics[width=1\linewidth]{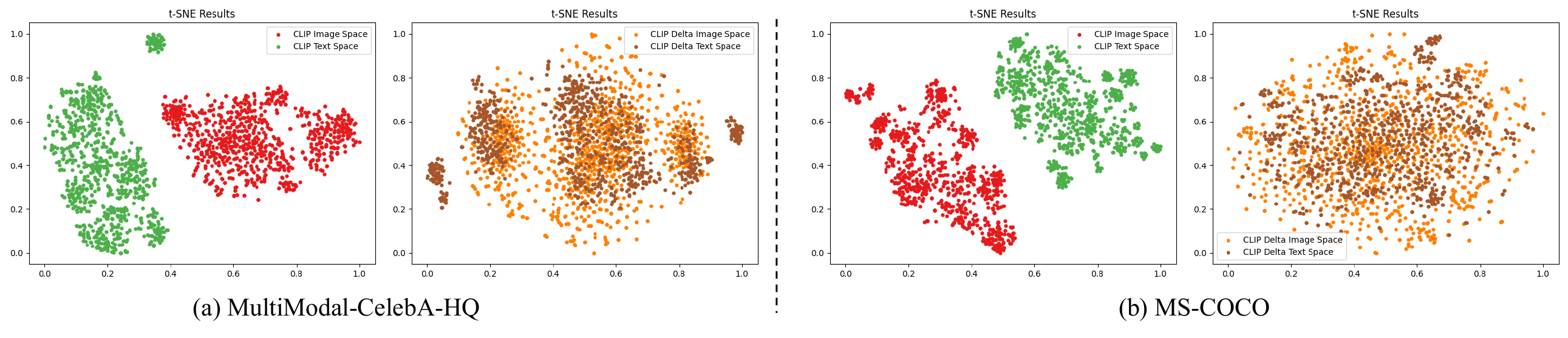}
    \caption{%
        Feature space analysis on multimodal datasets including (a) MultiModal-CelebA-HQ~\cite{xia2021tedigan} dataset and (b) MS-COCO~\cite{lin2014microsoft,chen2015microsoft} dataset. Paired CLIP image-text features (marked in red and green) and paired CLIP delta image-text features (marked in orange and brown) are visualized in 2D using t-SNE visualization.
    }
    \vspace{-3px}
    \label{fig:tsne}
\end{figure*}

Existing approaches~\cite{dong2017semantic,nam2018text,liu2020describe,li2020manigan,xia2021tedigan} typically train text-conditioned generative models from scratch with a large number of manually annotated image-text pairs. 
However, this process is very time-consuming as it requires expensive labor annotation.
Recently, the emergence of large-scale pre-trained vision language models, particularly Contrastive Language-Image Pre-training (CLIP) model~\cite{radford2021learning}, has brought new inspirations to this challenge. After trained on 400 million image-text pairs data, CLIP can embed real-world images and texts into a semantically aligned feature space. 
Leveraging this powerful property, CLIP-based methods~\cite{patashnik2021styleclip,liu2021fusedream,kocasari2022stylemc,wei2021hairclip,zhou2021lafite} have been proposed to avoid data collection and improve training efficiency with the CLIP prior.
Unfortunately, given one text prompt, these methods either leverage iterative optimization~\cite{xia2021tedigan,patashnik2021styleclip,liu2021fusedream}, or learn a specific mapping network~\cite{patashnik2021styleclip,gal2021stylegan,xu2022predict}, or manually tune hyper-parameters online~\cite{patashnik2021styleclip} to identify the fine-grained editing directions. Namely, \textbf{for different text prompts, they require different optimization processes}, which limits their flexibility during training or inference, result in poor generalization to unseen text prompts. 
So how to facilitate flexible text-guided image editing while taking full advantage of the CLIP prior?

Considering the meaningful semantic representation of the latent space within GANs (like pre-trained StyleGAN~\cite{karras2021alias,karras2019style,karras2020analyzing}), we aim to establish a comprehensive mapping from the CLIP feature space to this latent space. 

To achieve this, a straightforward solution is that we learn the editing direction of the latent space from CLIP image feature space, then perform the inference from the CLIP text feature space. 
When the image dataset has sufficient size, the model can be generalized to different text prompts, since CLIP provides a shared feature space between image and text modalities. 
However, in our empirical study we find there still exists modality gap between the image and text feature spaces~\cite{liang2022mind}, as illustrated in Fig.~\ref{fig:tsne}. It can be seen that the CLIP image space and the text space are not close to each other, while the CLIP feature difference space for image and text exhibits better alignment and semantic consistency. The former can be attributed to no exact one-to-one mapping between CLIP image and text space~\cite{patashnik2021styleclip}, and the latter indicates the CLIP feature direction of paired visual-textual data share similar semantic information (since the direction of CLIP features is semantically meaningful when all features are normalized to a unit norm~\cite{patashnik2021styleclip,zhou2021lafite}).
We refer to the CLIP feature difference space as CLIP DeltaSpace. 
Replacing the CLIP original feature space with the CLIP DeltaSpace has the potential to enable more precise and disentangled edits for the text-free trained image editing model.

\begin{figure*}[t]
\centering
\includegraphics[width=1\textwidth]{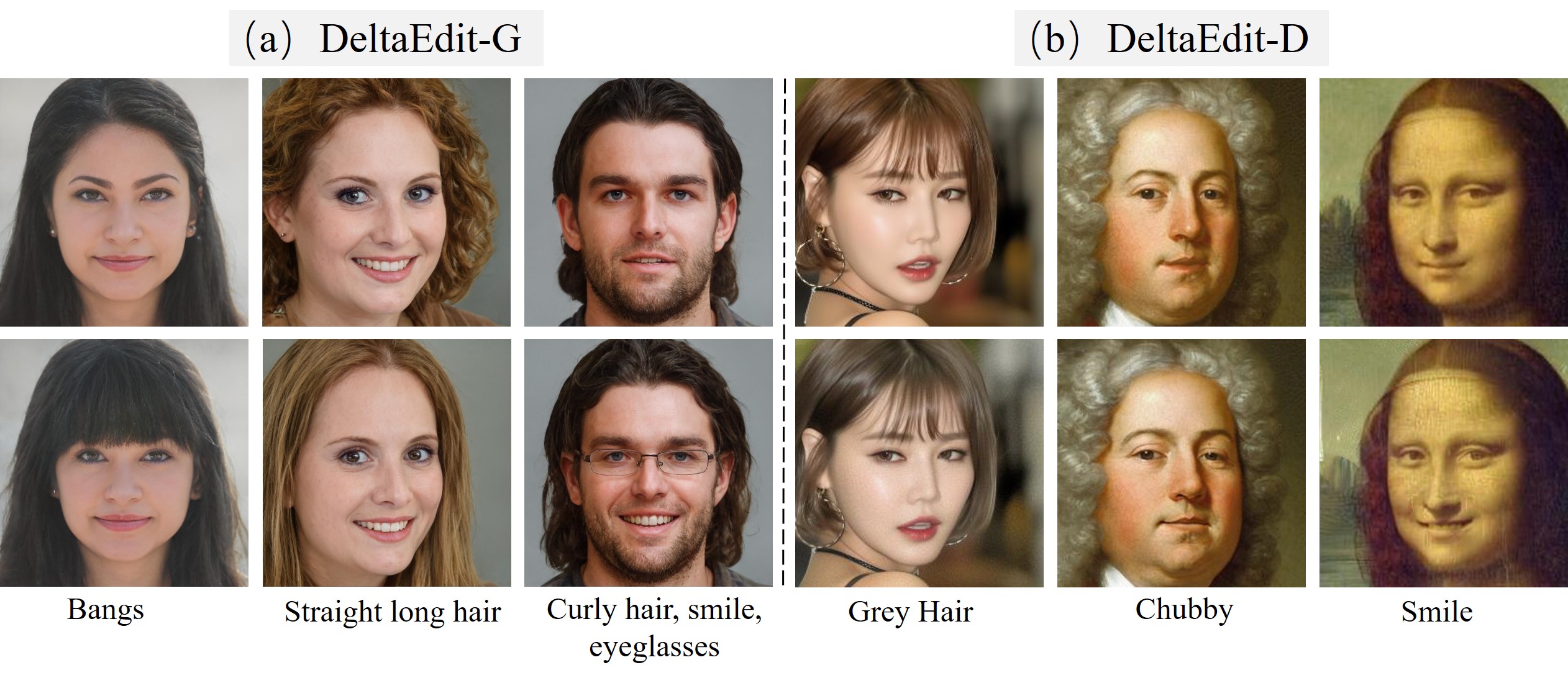}
\caption{Examples of text-guided image editing enabled by our DeltaEdit applied to the GAN model (named DeltaEdit-G) and the diffusion model (named DeltaEdit-D), respectively.}\label{fig:shouye}
\end{figure*}

Based on the above analysis, we propose a novel DeltaEdit framework. Specifically, our approach adopts a coarse-to-fine mapping network to learn the latent space direction of generative models from the CLIP image feature differences of randomly selected two images. In the inference stage, we apply the learned mapping onto the CLIP text feature of two given text prompts (one source prompt and one target prompt) to obtain the corresponding editing direction in the latent space, which is able to change image attributes from the source prompt to the target prompt after fed into the generative models. We employ the StyleGAN Style space~\cite{wu2021stylespace} as the latent space, which has been demonstrated to be semantically rich, interpretable, and disentangled. Besides, DeltaEdit is capable of controlling different types of generative models, including the GAN model and the diffusion model. As shown in Fig.~\ref{fig:shouye}, we present the instantiation results of DeltaEdit on both the GAN model and the diffusion model, showing that DeltaEdit can generalize well to various target text-prompts without bells and whistles.



The preliminary result of this work is published at~\cite{lyu2023deltaedit}. Compared to the conference paper, we include the following new contents:
(1) We provide a detailed analysis of DeltaSpace in Sec.~\ref{sec:deltaspace}. Based on the semantic-aligned properties of DeltaSpace, our DeltaEdit framework establishes effective, generalized and versatile editing performance on two types of generative models. The previous approach is only effective for the GAN model. 
(2) We propose a style-conditioned diffusion model in Sec.~\ref{sec:stylediffusion}, which leverages the semantic space of StyleGAN to control forward and reverse processes of the conditional diffusion model. This combination achieves strong detailed reconstruction while yielding remarkable image editing, as shown in Fig.~\ref{fig:shouye}.
(3) We include evaluations and comparisons of DeltaEdit-D to further validate the effectiveness and generalizability of the method in Sec.~\ref{sec:res_d} and Sec.~\ref{sec:com_d}, including semantically meaningful latent interpolation, real image reconstruction, flexible text-guided editing, \textit{etc}.
\section{Related Work}\label{sec:relate}

\subsection{Vision-language Representations}

Learning a shared and aligned feature space for generic Vision-language (VL) representation is of great importance. Following the success of BERT~\cite{devlin2018bert}, numerous large-scale pre-trained Vision-language (VL) models~\cite{li2020oscar,lu2019vilbert,su2019vl,tan2019lxmert,zhang2021vinvl} have been proposed and applied to various downstream tasks, including visual commonsense reasoning~\cite{zellers2019recognition}, image captioning~\cite{agrawal2019nocaps,lin2014microsoft}, and visual question answering~\cite{antol2015vqa,hudson2019gqa}.

A recent development, CLIP~\cite{radford2021learning}, has emerged as a powerful approach for joint vision-language learning. Trained on a vast dataset of 400 million (image, text) pairs, it strives to learn a joint semantical space for both images and texts using contrastive loss. Benefiting from its excellent image/text representation ability, CLIP has found extensive applications in different areas, such as domain adaptation~\cite{ge2022domain}, image segmentation~\cite{zhou2021denseclip,rao2021denseclip}, image generation and editing~\cite{gal2021stylegan,xu2022predict,li2022stylet2i,lyu2021sogan,sun2022anyface, lyu2023dran,li2023reganie,ma2023freestyle}. In particular, LAFITE~\cite{zhou2022towards} and KNN-Diffusion~\cite{sheynin2022knn} utilize the CLIP image-text feature space and exploit language-free text-to-image generation. 
Some researchers address the modality gap~\cite{liang2022mind,zhou2022lafite2}, explaining that CLIP features from different encoders concentrate on different narrow cones within the feature space, leading to inadequate alignment in the multi-modal feature space. However, for image editing task that require precise control, achieving better alignment within the multi-modal feature space is critical. We empirically find that the CLIP difference space (refered to as CLIP DeltaSpace) is better aligned than the original CLIP space. Within this well-aligned space, we achieve more precise and disentangled image editing.

\subsection{Text-guided Image Editing with GANs}

Text-guided image editing~\cite{dong2017semantic,nam2018text,liu2020describe,li2020manigan,xia2021tedigan,xu2022predict,kim2022diffusionclip} aims to manipulate the input images by using texts describing desired visual attributes (\textit{e.g.}, gender, age). Most previous methods utilize generative adversarial networks~(GANs)~\cite{goodfellow2014generative} to achieve image editing.

ManiGAN~\cite{li2020manigan} proposes affine combination module (ACM) and detail correction module (DCM) to generate new attributes matching the given text.  
Recently, some works~\cite{stap2020conditional,xia2021tedigan} have adopted StyleGAN as their backbone to perform image editing tasks. TediGAN~\cite{xia2021tedigan} aligns two modalities in the latent space of pre-trained StyleGAN by the proposed visual-linguistic similarity module. 
Furthermore, it only changes the attribute-specific layers when performing editing. 
More recently, StyleCLIP~\cite{patashnik2021styleclip} combines the generation power of StyleGAN and the image-text representation ability of CLIP~\cite{radford2021learning} to discover editing direction. 
They outline three approaches, namely Latent Optimization, Latent Mapper and Global Directions, which are denoted as StyleCLIP-OP, StyleCLIP-LM, and StyleCLIP-GD in this paper. 
The StyleCLIP-OP and StyleCLIP-LM are per-prompt training methods, which require optimizing the latent code or training a separate model for each text prompt. 
The third approach, StyleCLIP-GD, is a per-prompt fine-tuning method. It first finds global directions of semantic changes in StyleGAN's style space $\mathcal{S}$ by pre-defined relevance matrix of each channel in $\mathcal{S}$ to the image-space changes. During inference, it needs to manually tune hyper-parameters to discover the fine-grained directions for each specific text prompt.
%
In addition, StyleMC~\cite{kocasari2022stylemc}, another per-prompt fine-tuning method, proposes to find stable global directions with a combination of a CLIP loss and an identity loss. HairCLIP~\cite{wei2021hairclip} focuses more on the hair editing with the help of CLIP. FFCLIP~\cite{zhu2022one} collects 44 text prompts for face images to achieve facial attributes editing. 
FEAT~\cite{hou2022feat} focuses on enhancing edited regions by incorporating learned attention masks.
To improve the flexibility of text-guided image editing, we propose a novel framework to achieve various editing within a single model and do not require complex manual tuning. 

\subsection{Text-guided Image Editing with Diffusion Models}

Recently, diffusion models~\cite{sohl2015deep,dhariwal2021diffusion} have demonstrated the capability to generate high-quality images that rival those produced by GANs, while avoiding the instability associated with adversarial training. 
In previous work, Song \textit{et al.}~\cite{song2020score} propose score-based generative models for modeling data distributions using gradients. Ho \textit{et al.}~\cite{ho2020denoising} propose denoising diffusion probabilistic models (DDPMs) that achieve impressive sample quality through a combination of score-based generative models and diffusion models. Contrasting to the Markovian noise-injecting forward diffusion process assumed by DDPMs, Denoising Diffusion Implicit Models (DDIMs)~\cite{song2020denoising} adopt a non-Markovian forward process that had the same marginal distribution as DDPMs. DDIMs leverage the reverse denoising process associated with this non-Markovian process for sampling, effectively accelerating the sampling process. Moreover, DDIMs employ a deterministic forward-backward process, enabling nearly-perfect reconstruction, a capability that eludes DDPMs. Inspired by these seminal works, subsequent research works have focused on enhancing diffusion models in terms of sampling speed, sampling quality, and conditional synthesis. 

Unlike GANs~\cite{goodfellow2014generative}, diffusion models maintain a latent space that preserves the same dimensionality as the input.
Image editing in DiffusionCLIP~\cite{kim2022diffusionclip} is achieved by directly optimizing latent variables within the latent space of diffusion models under the supervision of CLIP. 
Diffusion Autoencoders~(DiffAE)~\cite{preechakul2022diffusion} introduces an additional encoder to capture a meaningful and decodable latent space for diffusion models. Moreover, Hierarchical Diffusion Autoencoders (HDAE)~\cite{lu2023hierarchical} leverages the coarse-to-fine feature hierarchy of the semantic encoder to encode semantic details. Meanwhile, Asyrp~\cite{kwon2022diffusion} selects the deepest feature maps of the pre-trained diffusion models in the U-Net~\cite{ronneberger2015u} architecture to represent semantics. 

In this work, we propose a novel diffusion model, named style-conditioned diffusion model, which utilizes the semantically rich and comprehensive representation offered by StyleGAN as the conditional signal to control the model.
\section{Methodology}\label{sec:method}

\subsection{DeltaSpace}\label{sec:deltaspace}


A semantic-aligned CLIP image-text feature space is crucial for flexible text-guided image editing, as it allows editing models to utilize image features as pseudo text features for text-free training. After trained on a sufficiently large image dataset, it can well generalize to various text prompts for zero-shot inference.

In this context, we empirically explore and identify a specific space with alignment properties, which we refer to as CLIP DeltaSpace. In this space, the CLIP visual feature differences between two images, denoted as $\Delta i$, and their corresponding CLIP text feature differences, denoted as $\Delta t$, are found to be semantically aligned. In other words, these two features exhibit high cosine similarity after normalization. This alignment indicates the direction in CLIP image feature space ($\Delta i$) keeps the similar semantic information as that in CLIP text feature space ($\Delta t$) \cite{patashnik2021styleclip,zhou2021lafite}.



\begin{figure}[t]
    \centering
    \includegraphics[width=1\linewidth]{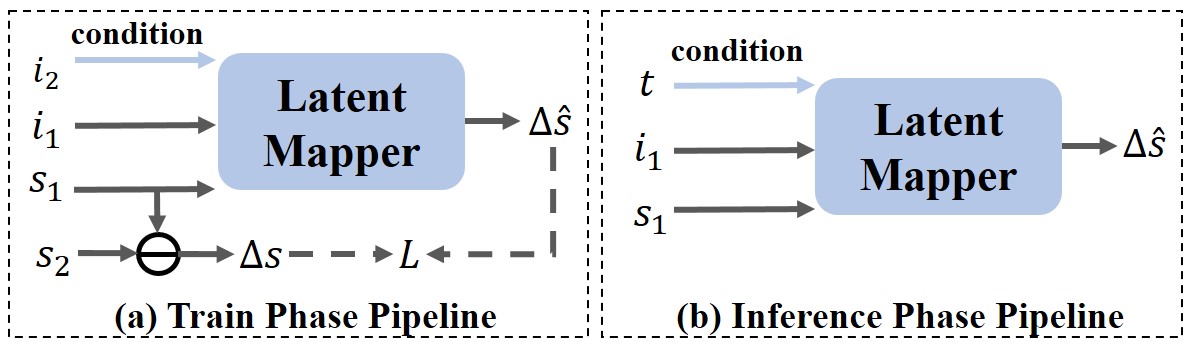}
    \caption{%
        Illustration of the straightforward solution to text-free training.
    }
    \label{fig:naive_framework}
\end{figure}

\begin{figure}[t]
    \centering
    \includegraphics[width=1\linewidth]{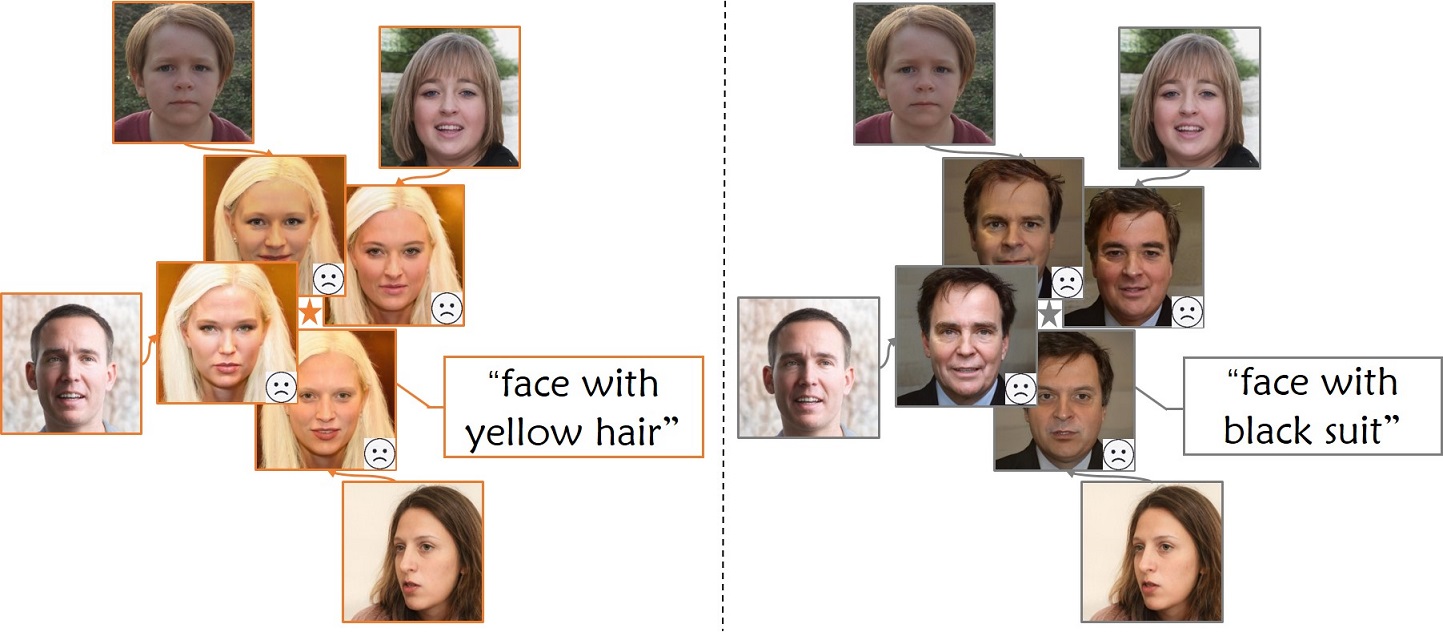}
    \caption{%
        The editing results of the straightforward solution. Take different source images as input, the method fits them all to an average face corresponding to the text-related attributes, by directly replacing the condition from image feature $i$ to text feature $t$.
    }
    \label{fig:naive}
\end{figure}

\subsubsection{Discussion with the Original CLIP Space}

To explore alignment and semantic continuity within the DeltaSpace, we conduct a comparative analysis with the original CLIP image-text space. The original CLIP space also exhibits alignment properties through contrastive learning on large-scale image-text pair data.

\begin{figure*}[t]
    \centering
    \includegraphics[width=1\linewidth]{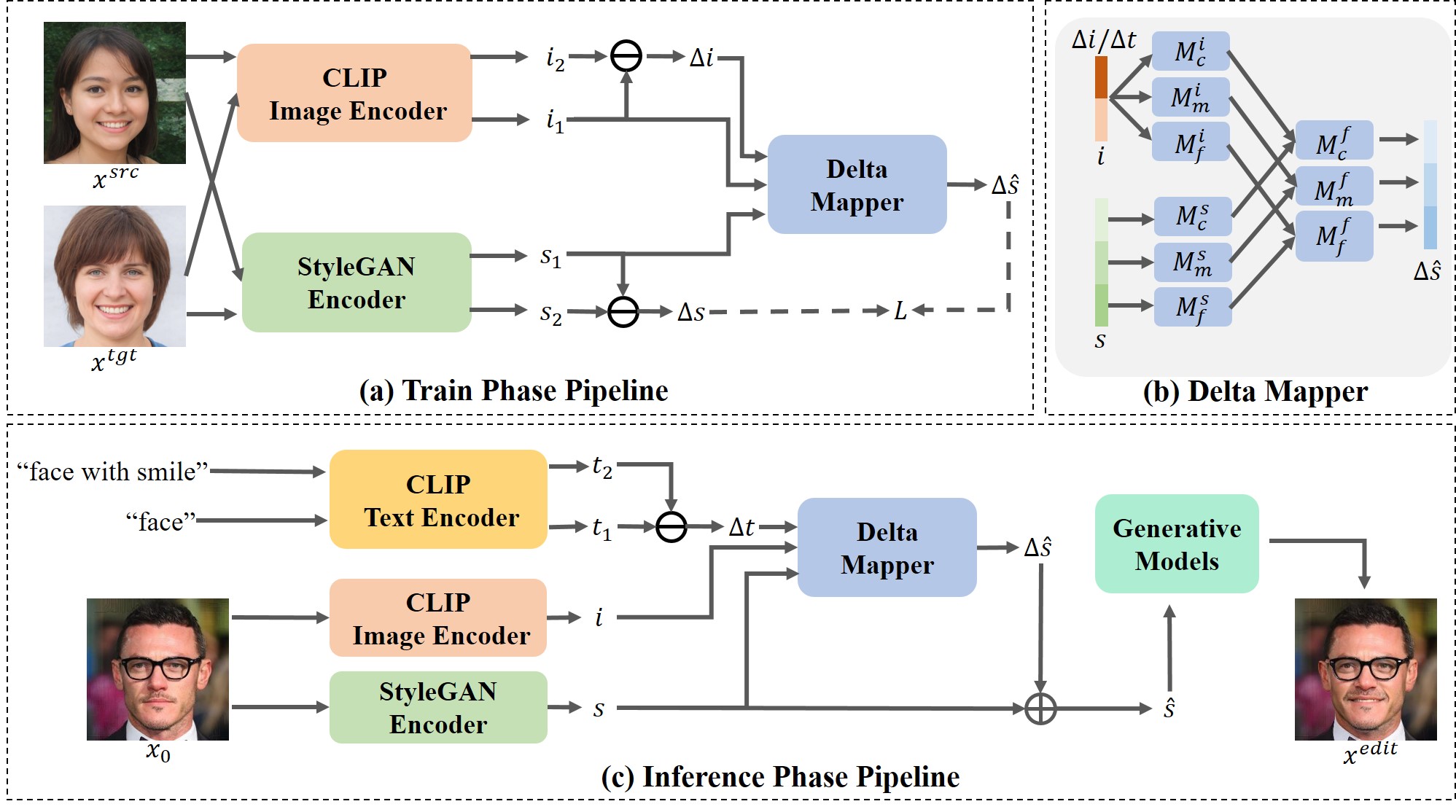}
    \caption{%
        The overall framework of the proposed DeltaEdit.
        (a) In the text-free training phase, we extract embeddings of two randomly selected images on CLIP image space and StyleGAN $\mathcal{S}$ space. Then we feed $i_1$, $s_1$ and $\Delta i$ into Delta Mapper to predict editing direction $\Delta \hat s$, which is supervised by $\Delta s$. 
        (b) The detailed architecture of the Delta Mapper, which achieves coarse-to-fine editing in three levels.
        (c) In the inference phase, based on the co-linearity between $\Delta i$ and $\Delta t$ in CLIP joint space, DeltaEdit can achieve text-guided image editing by taking two text prompts (denoting the source and desired target) as input.
    }
    \label{fig:framework}
\end{figure*}

We begin by employing t-SNE visualization on both spaces using two image-text datasets. As shown in Fig.~\ref{fig:tsne}, there is a noticeable modality gap between the CLIP image space and the CLIP text space, indicating that these two spaces are not closely related. In contrast, the CLIP DeltaSpace demonstrates better alignment and semantic consistency. Additionally, recent works~\cite{liang2022mind,sheynin2022knn,zhou2022lafite2} have also demonstrated that the original CLIP multi-modal space is not well-aligned due to model initialization and contrastive representation learning~\cite{liang2022mind}. 

We further propose a straightforward text-free editing method in the original CLIP space. Specifically, given two randomly selected images from the training dataset, one as the source image $x^{src}$ and the other as the target image $x^{tgt}$, we extract their CLIP image embeddings, denoted as $i_1$ and $i_2$. Additionally, we use a pre-trained StyleGAN inversion model~\cite{tov2021designing} as the encoder to extract their latent codes $s_1$ and $s_2$ in $\mathcal{S}$ space. Taking these extracted codes, we predict the editing direction $\Delta s = s_2 - s_1$.

As illustrated in Fig.~\ref{fig:naive_framework}, in the training phase, the source embeddings, $i_1$ and $s_1$, are taken as the input. The condition is from a target CLIP image embedding $i_2$ instead of a target CLIP text embedding. Then, the source embeddings and the image condition are sent to a latent mapper network to predict the editing direction $\Delta \hat{s}$:
\begin{equation}
    \Delta \hat s = LatentMapper(s_1, i_1, i_2)\text{.}
\end{equation}
During inference, we can predict the editing direction $\Delta \hat s$ as:
\begin{equation}
    \Delta \hat s = LatentMapper(s_1, i_1, t)\text{,}
\end{equation}
where $t$ is a CLIP text embedding constructed from a target text prompt.

However, as shown in Fig.~\ref{fig:naive}, given different source images and one target prompt, the straightforward way maps them all to average faces corresponding to the text-related attributes, leading to editing failures. This may be due to large differences in the CLIP image-text space, leading to inconsistencies in the training and testing.

Quantitative experiments in Tab.~\ref{tab:quatita} also demonstrate that the straightforward way cannot ensure accurate editing. 
In contrast, our proposed DeltaSpace allows for accurate and flexible image editing, as shown in Fig.~\ref{fig:framework}, demonstrating again DeltaSpace is more aligned than the original CLIP space.

\subsection{DeltaEdit}\label{sec:solution2}
Based on the semantic-aligned CLIP DeltaSpace, we introduce our DeltaEdit framework, which aims to establish a mapping network that bridges the DeltaSpace with the latent space direction of given generative models without any text supervision. 
The latent space we employ in DeltaEdit is the StyleGAN $\mathcal{S}$ space, which has rich semantics and superior disentanglement compared to other intermediate latent spaces~\cite{wu2021stylespace}, enabling precise control of target image attributes through editing directions in the $\mathcal{S}$ space. 

\noindent\textbf{Overview.} During training, we first take the extracted codes to get the CLIP image space direction $\Delta i = i_2 - i_1$ and StyleGAN $\mathcal{S}$ space direction $\Delta s=s_2 -s_1$. Then, as shown in Fig.~\ref{fig:framework}, we propose a latent mapper network, called Delta Mapper, to predict the editing direction as:
\begin{equation}
    \Delta \hat s = DeltaMapper(s_1, i_1, \Delta i)\text{,}
\end{equation}
where $s_1$ and $i_1$ are used as the input of Delta Mapper to provide specialized information for the source image.

During inference, we can achieve text-guided image editing with the trained Delta Mapper. As shown in Fig.~\ref{fig:framework} (c), given the source image $I$, we first extract its CLIP image embedding $i$ and StyleGAN $\mathcal{S}$ space embedding $s$. Then, we construct $\Delta t$ with source and target text prompts, and we can predict the editing direction $\Delta \hat s$ as:
\begin{equation}
    \Delta \hat s = DeltaMapper(s, i, \Delta t),
\end{equation}
where the editing direction is subsequently used to generate edited latent embedding $\hat s = s + \Delta \hat s$. In the final step, we can generate the synthesized image $x^{edit}$ with a given generative model conditioned on $\hat s$.

\noindent\textbf{Delta Mapper.}
The architecture of the Delta Mapper is illustrated in Fig.~\ref{fig:framework} (b).
Since StyleGAN has the property that different layers correspond to different semantic levels~\cite{karras2019style,xia2021tedigan}, it is common to divide these layers into different levels and implement coarse-to-fine editing within each level. Following \cite{patashnik2021styleclip,wei2021hairclip}, we adopt three levels of sub-modules~(coarse, medium, and fine) for each designed module. Each sub-module contains several fully-connected layers.
For different levels of the source latent code $s_1$, we first propose \textit{Style Module} to obtain coarse-to-fine intermediate features $(e^s_c, e^s_m, e^s_f)$, where subscripts stand for coarse, medium, and fine levels and superscripts stand for the Style Module. Then, we concatenate the $\Delta i$ and $i_1$ as input, and propose \textit{Condition Module} to learn coarse, medium and fine embeddings $(e^i_c, e^i_m, e^i_f)$ separately, which have the same dimensions as $(e^s_c, e^s_m, e^s_f)$.
In the final step, we fuse generated coarse-to-fine features using proposed \textit{Fusion Module} with three sub-modules ($M^f_c(\cdot,\cdot), M^f_m(\cdot,\cdot), M^f_f(\cdot,\cdot)$) and predict editing direction as:
\begin{equation}
\begin{aligned}
    \Delta \hat s = &(M^f_c(e^i_c, e^s_c), M^f_m(e^i_m, e^s_m), M^f_f(e^i_f, e^s_f)). 
\end{aligned}
\end{equation}

To train the proposed Delta Mapper, our full objective function contains two losses, which can be denoted as:
\begin{equation}
\begin{aligned}
    \mathcal{L}\!=\!\mathcal{L}_{rec}\!+\!\mathcal{L}_{sim}\!=\! \|\Delta \hat s\!-\!\Delta s\|_2 +\!1\!- cos(\Delta \hat s,\Delta s),
\end{aligned}
\end{equation}
where $\mathcal{L}$-2 distance reconstruction loss is utilized to add supervision for learning the editing direction $\Delta \hat s$ and cosine similarity loss is introduced to explicitly encourage the network to minimize the cosine distance between the predicted embedding direction $\Delta \hat s$ and $\Delta s$ in the $\mathcal{S}$ space.

\noindent\textbf{Text-prompt Setting.}
In the inference phase, the crucial issue is how to construct $\Delta t = t_2 - t_1$. To be consistent with the training phase, the source text and target text should be specific text descriptions of two different images. However, the DeltaEdit network will generate images towards the direction of the target text and the opposite direction of the source text in this manner. Therefore, to remove the reverse effect of the source text prompt containing attributes, we naturally place all the user-described attributes in the target text prompts. Take human portrait images as an example, if the user intends to add ``smile'' to a face, then our DeltaEdit uses ``face with smile'' as the target text and ``face'' as the source text. An advantage is that the importance of words that indicate the desired attributes (like ``smile'') can be enhanced, which alleviates the problem that CLIP is not sensitive to fine-grained or complex words~\cite{radford2021learning,saharia2022photorealistic}.) This text-prompt setting is also applicable to other editing domains. In experiments, we further verify that it enables accurate image editing.

\noindent\textbf{Disentanglement.}
To improve the disentanglement, we further optimize the obtained editing direction $\Delta \hat s$ using the pre-computed relevance matrix $R_s$, which records how CLIP image embedding changes when modifying each dimension in $\mathcal{S}$ space ~\cite{patashnik2021styleclip}. We can set some channels of $\Delta \hat s$ as zero if the channels have a low correlation to the target text according to $R_s$.


%

\section{Instantiation of DeltaEdit}\label{sec:solution3}

To demonstrate the effectiveness of our work, we instantiate the trained DeltaEdit framework on both the GAN model (named DeltaEdit-G) and the diffusion model (named DeltaEdit-D).

\subsection{DeltaEdit-G}

Once the bridge between the CLIP DeltaSpace and the $\mathcal{S}$ space is established, the Delta Mapper network can accurately predict the editing direction $\Delta \hat s$ in the $\mathcal{S}$ space, conditioned on the corresponding CLIP text embeddings $\Delta t$. 
For generating the edited images with target attributes, we leverage the pre-trained StyleGAN~\cite{karras2021alias,karras2019style,karras2020analyzing}, known for its high-quality generation capability, as the image decoder. The final edited image $x^{edit}$ could be formulated as following: 
\begin{equation}
    x^{edit} = G(s+\Delta \hat s). 
\end{equation}
%

\subsection{DeltaEdit-D}

In Sec.~\ref{sec:preliminaries}, we briefly overview the background knowledge about diffusion models. Then, in Sec.~\ref{sec:stylediffusion}, we present our proposed style-conditioned diffusion model and the instantiation of DeltaEdit for controlling the model and facilitating image editing.

\subsubsection{Preliminaries}\label{sec:preliminaries}


\noindent\textbf{Denoising Diffusion Probability Model (DDPM).}
DDPM~\cite{ho2020denoising,sohl2015deep} comprises a forward diffusion process and a reverse inference process.
%
The forward process is a Gaussian noise perturbation parameterized with a Markovian process:
\begin{equation}
    q\left(x_t \mid x_{t-1}\right)=\mathcal{N}\left(x_t ; \sqrt{1-\beta_t} x_{t-1}, \beta_t I\right)\text{.}
\label{eq:eq1}
\end{equation}
Here, $x_t$ is the noised image at time-step $t$, $\beta_t$ is a fixed or learned scale factor. The Eq.~\ref{eq:eq1} can be further simplified as:
\begin{equation}
    \label{eq:eq2}
    q\left(x_t \mid x_0\right)=\mathcal{N}\left(x_t ; \sqrt{\alpha_t} x_0, \left(1-\alpha_t\right) I\right)\text{.}
\end{equation}
where $\alpha_t=\prod_{s=1}^t\left(1-\beta_s\right)$.

The reverse process is parameterized as another Gaussian transition:
\begin{equation}
    p_{\theta}\left(x_{t-1} \mid x_{t}\right)=\mathcal{N}\left(x_{t-1}; \mu_{\theta}\left(x_t, t\right)x_{t}, \sigma^2_t I\right)\text{,}
\end{equation}
where $\mu_{\theta}$ can be expressed as the linear combination of $x_{t}$ and a noise predictor $\epsilon_\theta\left(x_{t}, t\right)$, which can be learned by minimizing the object:
\begin{equation}
    \mathbb{E}_{x_0 \sim q\left(x_0\right), \epsilon \sim \mathcal{N}(0, I), t}\left\|\epsilon-\epsilon_\theta\left(x_t, t\right)\right\|_2^2\text{.}
\end{equation}


\noindent\textbf{Denoising Diffusion Implicit Model (DDIM).}
DDIM is a non-Markovian noising process proposed by Song \textit{et al.}~\cite{song2020denoising} and it redefines Eq.~\ref{eq:eq1} as
\begin{equation}
	\begin{aligned}
	q_\sigma\left(x_{t-1} \mid x_t, x_0\right)&=\mathcal{N}(x_{t-1};\sqrt{\alpha_{t-1}} x_0 \\
	&+\sqrt{1-\alpha_{t-1}-\sigma_t^2} \cdot \frac{x_t-\sqrt{\alpha_t} x_0}{\sqrt{1-\alpha_t}}, \sigma_t^2 I) \text{.}
	\end{aligned}
\label{eq:eq6}
\end{equation}
The reverse process of DDIM is
\begin{equation}
	\begin{aligned}
    x_{t-1}&=\sqrt{\alpha_{t-1}} \underbrace{\left(\frac{x_t-\sqrt{1-\alpha_t} \epsilon_\theta\left(x_t, t\right)}{\sqrt{\alpha_t}}\right)}_{\text {``predicted } x_0 \text { " }}\\
	&+\underbrace{\sqrt{1-\alpha_{t-1}-\sigma_t^2} \cdot \epsilon_\theta\left(x_t, t\right)}_{\text {``direction pointing to } x_t \text{"}}+\underbrace{\sigma_t z_t}_{\text {random noise}}\text{,}
	\end{aligned}
\label{eq:7}
\end{equation}
where $z_t \sim \mathcal{N}(0, I)$. When $\sigma_t$ = 0, this process is deterministic and each original sample can be nearly perfectly reconstructed through the reverse process.


\begin{figure}[t]
    \centering
    \includegraphics[width=1\linewidth]{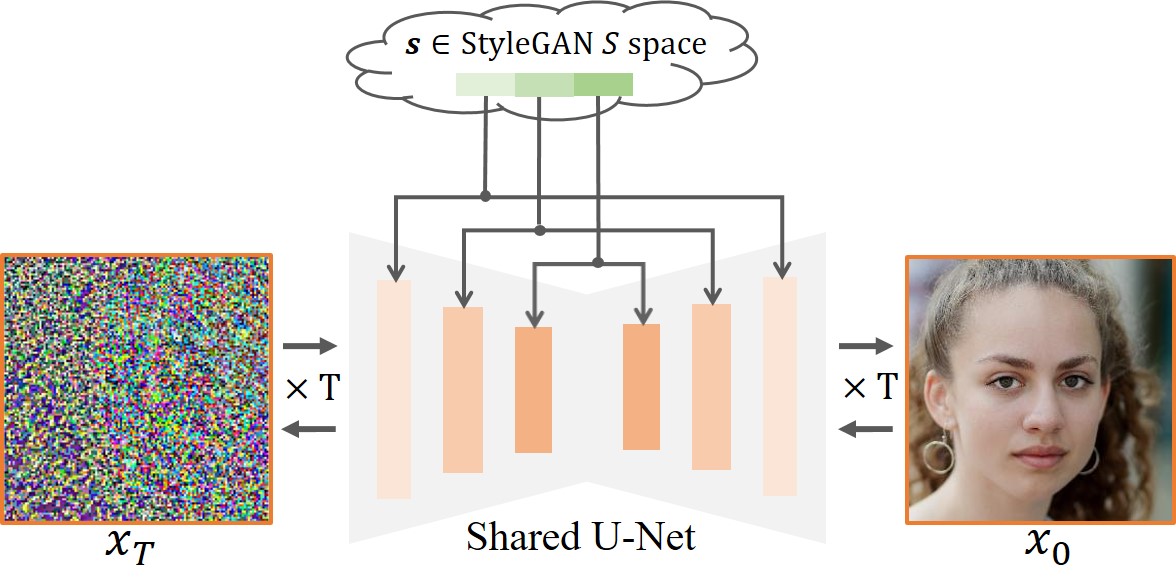}
    \caption{%
        Overview of our proposed style-conditioned diffusion model, which is based on StyleGAN $\mathcal{S}$ space. Here, the $s$ feature captures the high-level semantics while $x_T$ captures low-level stochastic variations.
    }
    \label{fig:diffusion}
\end{figure}

\subsubsection{Style-conditioned Diffusion}\label{sec:stylediffusion}

We present a style-conditioned diffusion model, which conditioned on semantic-rich Style space. Specifically, in the training phase, the diffusion model learns to predict noises based on the given image $x_0$ and corresponding style code $s$. In the inference phase, conditioned on the new style code $\hat{s}$ obtained from the trained DeltaMapper in Fig.~\ref{fig:framework}, the diffusion model can obtain the corresponding editing image $x^{edit}$ by stepwise denoising. 
The specific training and inference processes are described as follows.
%

\begin{figure*}[t]
    \centering
    \includegraphics[width=1\linewidth]{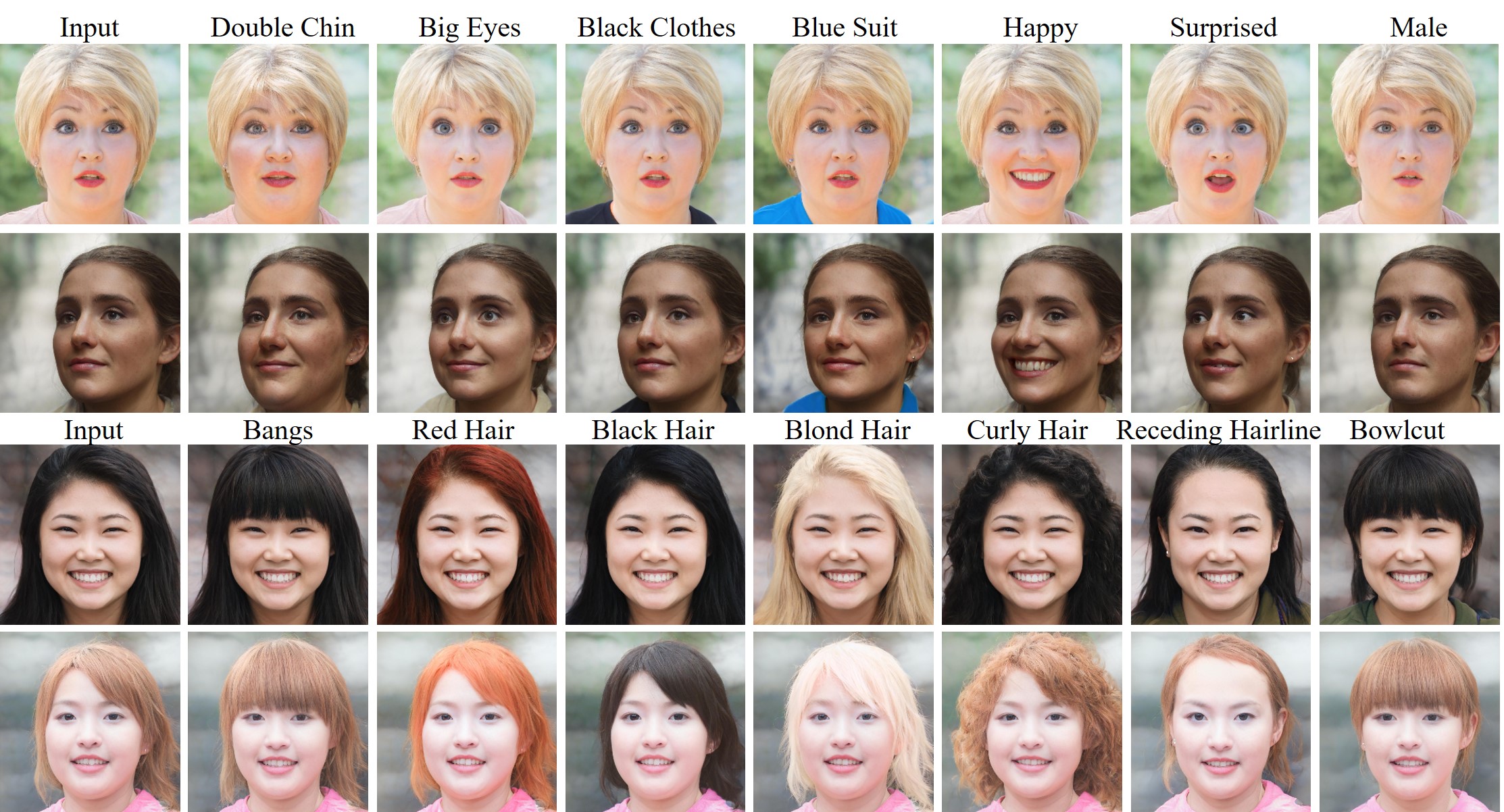}
    \caption{%
        Results of DeltaEdit-G for facial images on StyleGAN2 FFHQ model.
        The target attribute included in the text prompt is above each image.}
    \label{fig:face_conf}
\end{figure*}

During the training process, the style code $s$ of the original image $x_0$ is extracted from a GAN inversion network~\cite{tov2021designing} and divided into coarse, medium, and fine features, denoted as $(e^s_c, e^s_m, e^s_f)$. 
Then, these hierarchical features are utilized to modulate different levels of the U-Net network within the diffusion model. 
As shown in Fig.~\ref{fig:diffusion}, the features $h$ of the U-Net are also partitioned into coarse, medium and fine features, represented as $(e^h_c, e^h_m, e^h_f)$, based on their spatial dimensions.
Following previous works~\cite{dhariwal2021diffusion,preechakul2022diffusion,lu2023hierarchical}, we employ adaptive group normalization for hierarchical conditional encoding, denoted as $AdaGN(e^h_{l}, e^s_{l}, t)$, where semantic level $l \in \{c,m,f\}$.
This hierarchical conditional encoding enables precise semantic control and while aligning with the DeltaMapper architecture described in Sec.~\ref{sec:solution2}.
Additionally, to generate $x_T$, we utilize the deterministic forward process of DDIM while conditioning it on $s$ through the noise predictor $\epsilon_\theta$. To train $\epsilon_\theta$, the simple version of DDPM loss~\cite{ho2020denoising} is used:
\begin{equation}
    \label{eq:eq:9}
    \mathcal{L}_{\text {simple}}=\mathbb{E}_{x_0 \sim q\left(x_0\right), \epsilon_t \sim \mathcal{N}(0, I), t}\left\|\epsilon_\theta\left(x_t, t, s\right)-\epsilon_t\right\|_1\text{.}
\end{equation}
%
%
Note that $x_T$ is not required during training.


During the inference process, take obtained $s$ and $x_T$ as input, the trained diffusion acted as the decoder to generate the corresponding original image $x_0$. Specifically, it models $p_\theta(x_{t-1}\mid x_t, s)$ to match $q(x_{t-1} \mid x_t, x_0)$ in Eq.~\ref{eq:eq6}, with the following reverse process in a deterministic manner:
\begin{equation}
    \label{eq:our_ddim}
    p_\theta\left(x_{0:T} \mid s\right)=p\left(x_T\right) \prod_{t=1}^T p_\theta\left(x_{t-1} \mid x_t, s\right)\text{.}
\end{equation}
According to Song~\textit{et al.}~\cite{song2020denoising}, $p_\theta\left(x_{t-1} \mid x_t, s\right)$ can be parameterized as the noise predictor $\epsilon_\theta\left(x_{t}, t, s\right)$.


Finally, to achieve text-guided image editing with the trained diffusion, DeltaEdit framework is utilized to acquire the edited direction $\Delta \hat{s}$ from given text prompts. Subsequently, the edited style code is obtained by $\hat s = s + \Delta \hat s$.
With $(\hat s, x_T)$, we leverage the reverse DDIM process described in Eq.~\ref{eq:our_ddim} to produce the final edited image $x^{edit}$. 
\section{Experiments}\label{sec:exp}

\subsection{Experimental Setups}

\noindent\textbf{Datasets.}
To verify the effectiveness and generalization of the proposed method, we conduct extensive experiments in a range of challenging domains. For the face image domain, to construct robust Delta Mapper of our DeltaEdit framework, we randomly choose 58,000 images from FFHQ~\cite{karras2019style} dataset and sample 200,000 fake images from StyleGAN for training. We use the remaining 12,000 FFHQ~\cite{karras2019style} images and CelebA-HQ~\cite{karras2017progressive} images for evaluation. Similarly, for the proposed style-conditioned diffusion model, we train on FFHQ and test on FFHQ (intra-dataset) and CelebA-HQ (cross-dataset). We additionally provide results of DeltaEdit-G on the LSUN~\cite{yu2015lsun} Cat, Church and Horse datasets. Note that all real images are inverted by e4e encoder~\cite{tov2021designing} to obtain $s$ codes.

\begin{figure}[t]
    \centering
    \includegraphics[width=1\linewidth]{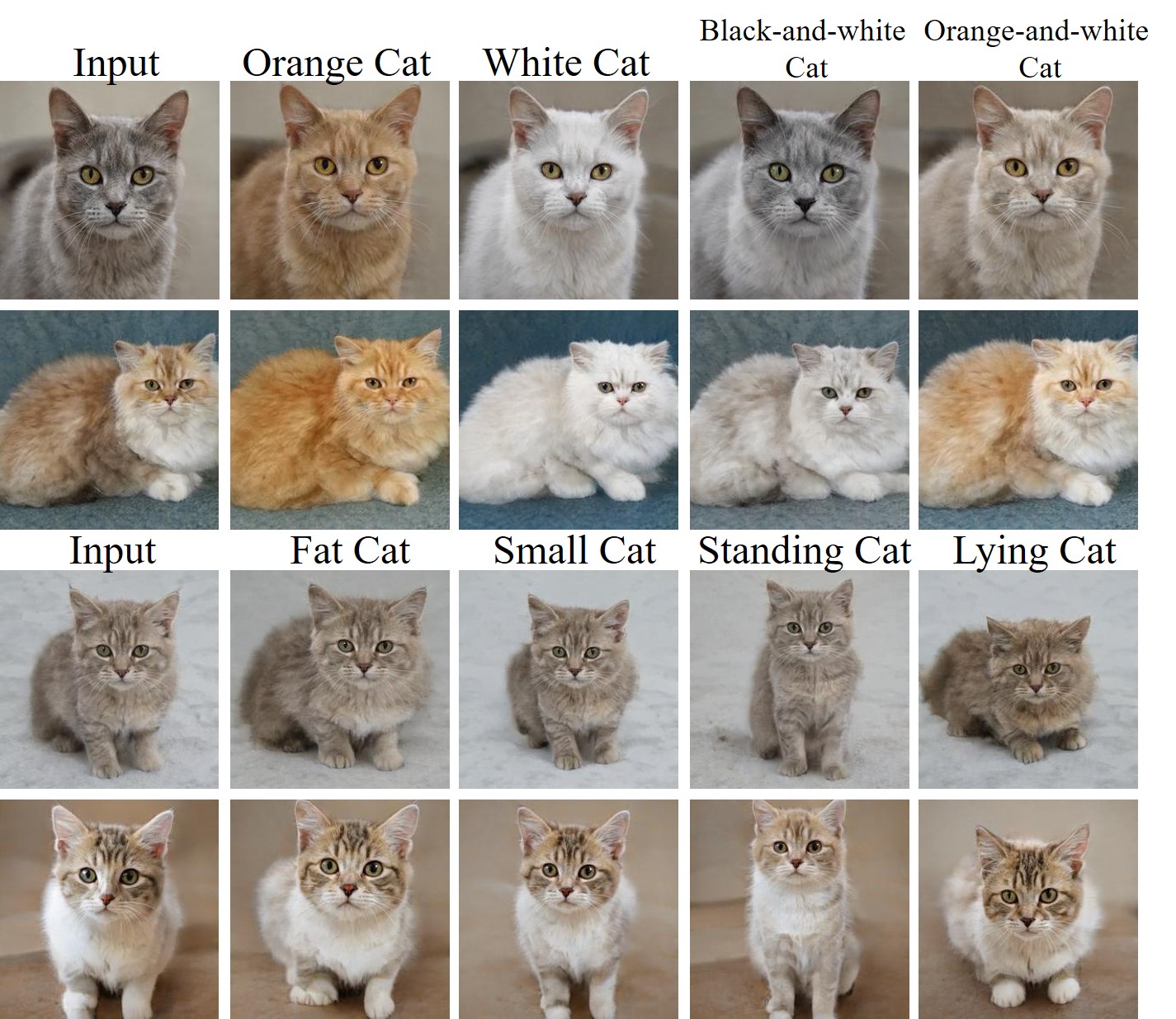}
    \caption{%
        Results of DeltaEdit-G for cat images, using StyleGAN2 pre-trained on LSUN cats dataset.
        The target text prompt is indicated above each column.
    }
    \label{fig:cat}
\end{figure}

\begin{figure*}[t]
    \centering
    \includegraphics[width=1\linewidth]{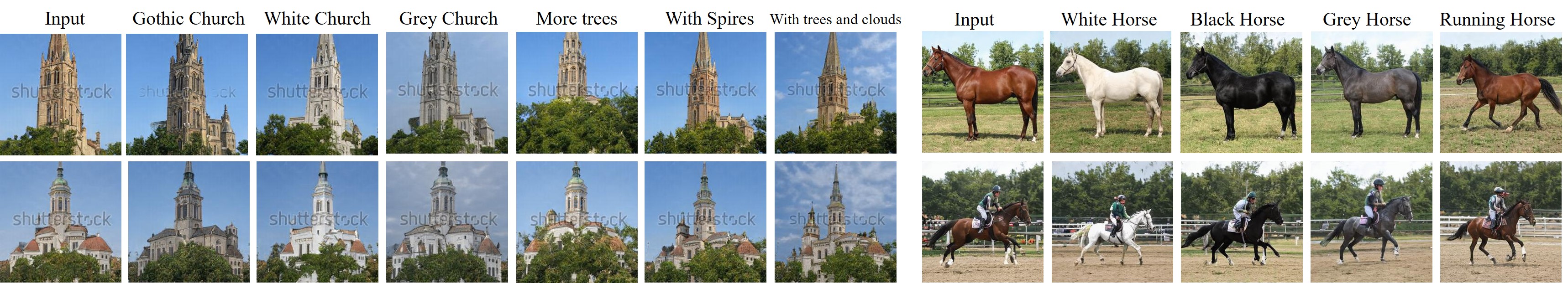}
    \caption{%
        Results of DeltaEdit-G for church and horse images, using StyleGAN2 pre-trained on LSUN churches dataset (left) and LSUN horses dataset (right) separately.
    }
    \label{fig:churchhorse}
\end{figure*}

\begin{figure*}[t]
    \centering
    \includegraphics[width=1\linewidth]{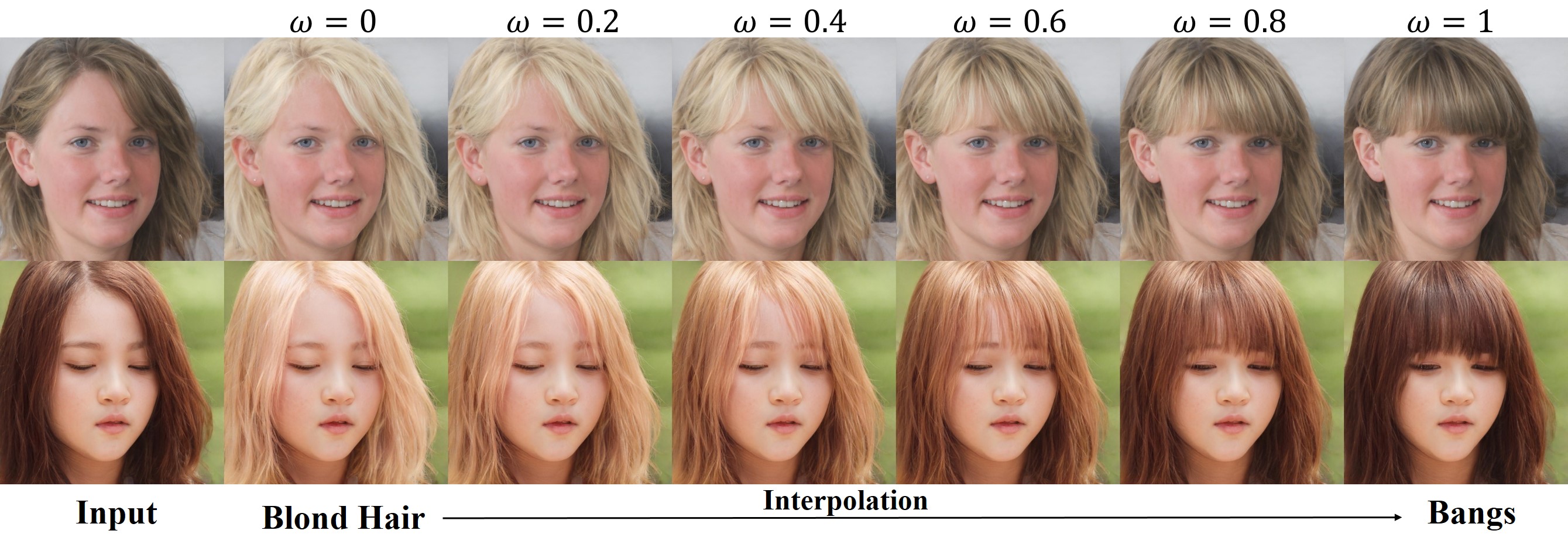}
    \caption{%
        Results of facial attribute interpolation. The styles of the interpolated images continuously transfer from attribute A~(``Blond Hair'') to attribute B~(``Bangs'') by setting the weight $\omega$ from 0 to 1 in intervals of 0.2.
    }
    \label{fig:inter}
\end{figure*}

\noindent\textbf{Compared Methods.}
In our comparisons, we evaluate our proposed method against GAN-based and diffusion-based methods. Among GAN-based methods, we include TediGAN~\cite{xia2021tedigan}, StyleCLIP-LM~\cite{patashnik2021styleclip}, StyleCLIP-GD~\cite{patashnik2021styleclip}, and StyleMC~\cite{kocasari2022stylemc}. For diffusion-based methods, we include DiffusionCLIP~\cite{kim2022diffusionclip}, DiffAE~\cite{preechakul2022diffusion}, Asyrp~\cite{kwon2022diffusion}, NTI~\cite{mokady2023null} and PTI~\cite{dong2023prompt}. For fair comparisons, we follow the official experimental settings of all compared methods.

\noindent\textbf{Implementation Details.}
The proposed Delta Mapper is trained on 1 NVIDIA Tesla P40 GPU. During training, we set a batch size of 64 and adopt the ADAM~\cite{kingma2014adam} optimizer with $\beta_1=0.9$, $\beta_2=0.999$, and a constant learning rate of 0.5. As for the proposed style-conditioned diffusion model, it is trained on 8 NVIDIA Tesla V100 GPUs. During training, we utilize a batch size of 64 and maintain a fixed learning rate of $1e^{-4}$. For the optimizer, we opt for AdamW~\cite{loshchilov2017decoupled} with a weight decay of 0.01, following DiffAE~\cite{preechakul2022diffusion}.

\noindent\textbf{Evaluation Metrics.}
To assess the image editing performance of various methods, we utilize three evaluation metrics: (1) \textit{Fréchet Inception Distance} (FID) ~\cite{heusel2017gans}, (2) \textit{Peak Signal-to-Noise Ratio} (PSNR) and (3) \textit{Identity Similarity before and after editing by Arcface}~\cite{deng2019arcface} (IDS). A lower FID value indicates better performance, whereas higher PSNR and IDS values indicate better image quality and better disentanglement capability, respectively. Following~\cite{preechakul2022diffusion,lu2023hierarchical}, to evaluate the image reconstruction performance of different methods, we choose evaluation metrics of (1) Structural Similarity Index (SSIM)~\cite{wang2004image}, (2) Learned Perceptual Image Patch Similarity (LPIPS)~\cite{zhang2018unreasonable}, and (3) Mean Squared Error~(MSE). Higher SSIM,and lower LPIPS, MSE represent fewer reconstruction errors.

\subsection{Results of DeltaEdit-G}

\subsubsection{Qualitative Results}

\noindent\textbf{Text-guided Image Editing.}
In Fig.~\ref{fig:face_conf}, we show facial editing results of 14 attributes on StyleGAN2 FFHQ model, which are generated from one trained DeltaEdit model. The results show that only the target attributes are manipulated, while other irrelevant attributes are well preserved. For example, we achieve accurate hairstyle editing while keeping other visual attributes that are not related to hair unchanged. Meanwhile, the results are well adapted to the individual with diverse details, rather than overfitting to the same color or shape, which can be seen in ``red hair'' and ``bowlcut hairstyle'' obviously.


%
For cat images, the manipulated results are shown in Fig.~\ref{fig:cat}, where StyleGAN2 pretrained on LSUN cats datasets~\cite{yu2015lsun} is used. As shown in the upper sub-figure in Fig.~\ref{fig:cat}, we can manipulate the fur color of input cats with almost no change in shape or posture. We can also perform editing driven by complex concepts, like ``black-and-white cat'' and ``orange-and-white cat''. As shown in the lower sub-figure in Fig.~\ref{fig:cat}, we can control over the movement of the input cat image by the text descriptions of ``standing cat'', ``lying cat'', \textit{etc}.
For church and horse images, a variety of manipulated results are provided in Fig.~\ref{fig:churchhorse}. All edits in Fig.~\ref{fig:churchhorse} are conducted using StyleGAN2 pretrained on LSUN churches and horses datasets~\cite{yu2015lsun} and all input images are generated images from StyleGAN2. By using the text prompts like ``more trees'', the corresponding attributes are manipulated in the generated churches. 

It is worth mentioning that all text prompts have never been seen during training, which further indicates the effectiveness of the proposed method.

\noindent\textbf{Attribute Interpolation.}
We present the attribute interpolation results of our approach. Given two edited latent codes $\hat s_a$ and  $\hat s_b$ in  $\mathcal{S}$ space, we calculate the new edited latent codes using the formula $s^{\omega} = \omega \cdot \hat s_a + (1-\omega)\cdot \hat s_b$, where $\omega$ is the interpolation weight. As depicted in Fig.~\ref{fig:inter}, the first row illustrates the results of our method applied to StyleGAN, while the second row shows the results when our approach is employed with the designed diffusion model. We successfully achieve natural and impressive interpolated results from reference text A~(``Blond Hair'') to reference text B~(``Bangs''), by adjusting the weight $\omega$ from 0 to 1 in intervals of 0.2.

\begin{figure}[t]
    \centering
    \includegraphics[width=1\linewidth]{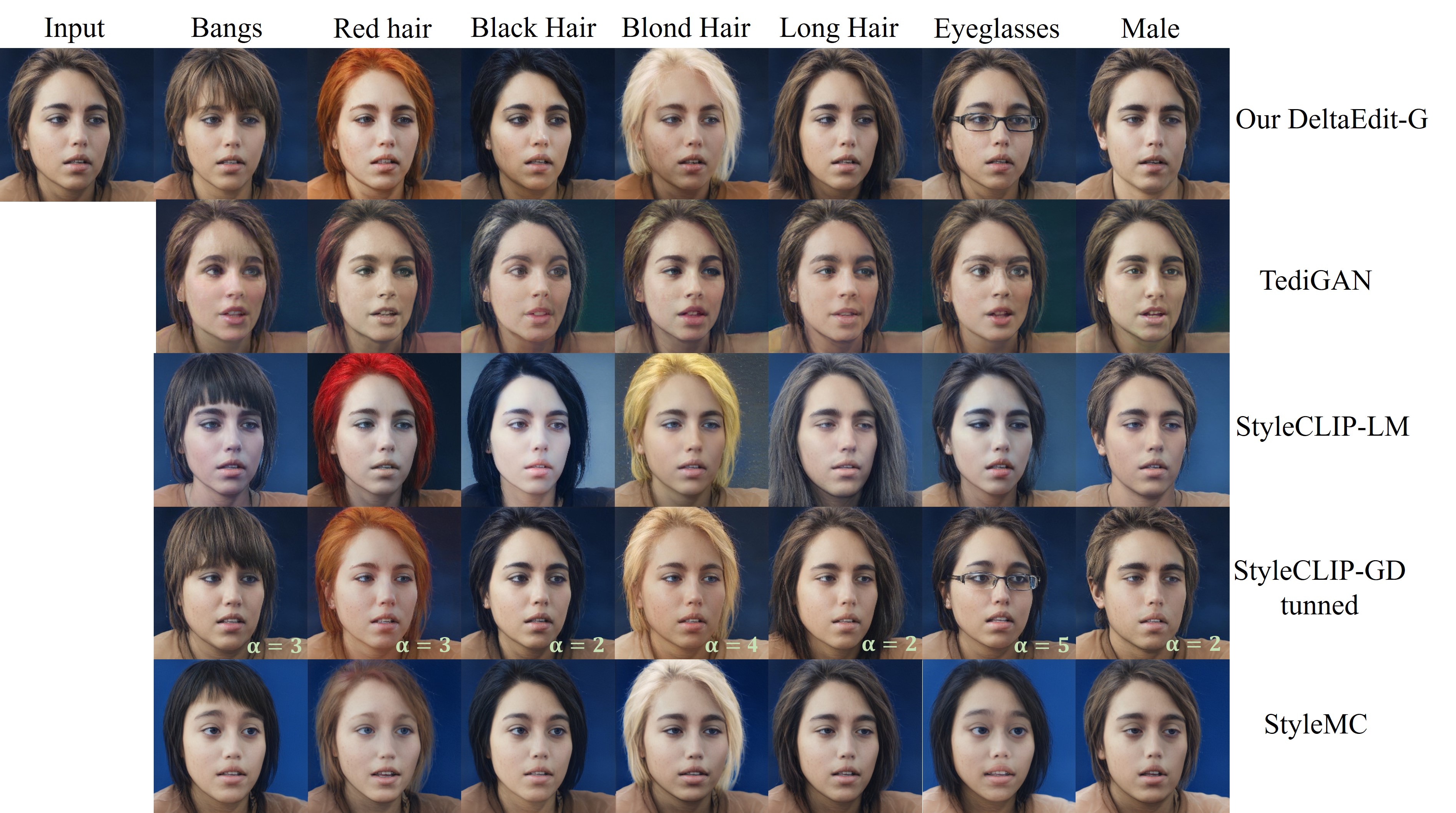}
    \caption{%
        Comparison results with GAN-based methods, including TediGAN, StyleCLIP-LM, StyleCLIP-GD and StyleMC. Our DeltaEdit-G approach demonstrates better visual realism and attribute disentanglement almost in all cases.
    }
    \label{fig:comwm}
\end{figure}

\begin{figure}[t]
    \centering
    \includegraphics[width=1\linewidth]{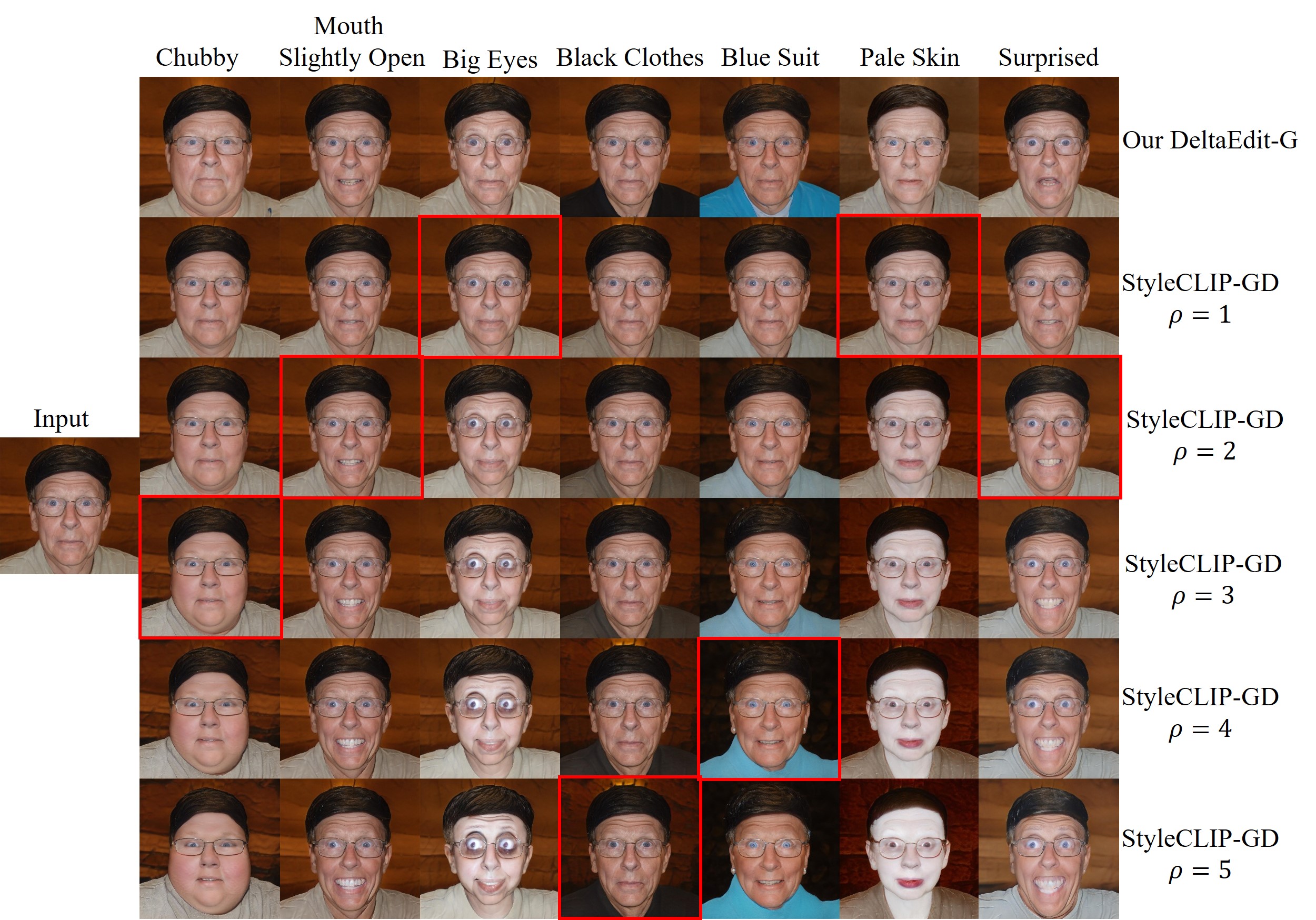}
    \caption{%
        The comparison between DeltaEdit-G and StyleCLIP-GD under different editing strengths $\rho$. Without complex hyper-parameters tuning, our method achieves more natural and disentangled results, compared to the best results of StyleCLIP-GD tuned under different parameters~(labeled with red boxes).
    }
    \label{fig:com}
\end{figure}

\subsubsection{Qualitative Comparison}

For a fair comparison, we compare DeltaEdit-G with state-of-the-art GAN-based image editing methods, TediGAN~\cite{xia2021tedigan}, StyleCLIP-LM~\cite{patashnik2021styleclip}, StyleCLIP-GD~\cite{patashnik2021styleclip} and StyleMC~\cite{kocasari2022stylemc}. 
As shown in Fig.~\ref{fig:comwm}, the results produced by TediGAN are highly entangled and almost fail in editing attributes, such as ``long hair'' and ``eyeglasses''. The results of StyleCLIP-LM and StyleMC are unstable and also entangled. 
Moreover, the results of StyleCLIP-GD have better disentanglement than StyleCLIP-LM as it is well-tunned for each case. However, the disadvantage of StyleCLIP-GD is the complex hyper-parameters tunning and long inference time as reported in Tab.~\ref{table:time}. In comparison, DeltaEdit-G yields the most impressive and disentangled results almost in all cases.

To verify the inference flexibility of our method, we further compare DeltaEdit-G with the strong baseline, \textit{i.e.}, StyleCLIP-GD, which achieves editing by manual tuning two hyper-parameters, disentanglement threshold $\tau$ and editing strength $\rho$.
We also introduce disentanglement threshold $\tau$ to control some channels of the obtained $\Delta \hat s$ as zero. For a fair comparison, we empirically set $\tau$ at 0.03 and compare with StyleCLIP-GD under different editing strength $\rho$ in Fig.~\ref{fig:com}.
The results demonstrate that StyleGAN-GD cannot generalize to different image attributes, \textit{i.e.}, it cannot produce natural editing results under the same hyper-parameters.
For example, the result conditioned on ``big eyes'' is the most accurate at $\rho=1$ while the result conditioned on ``black clothes'' is the most correct when $\rho=4$. In contrast, our method achieves ideal editing results of various text prompts without manually tuning the editing strength. 
Moreover, under the same disentanglement threshold $\tau$, StyleCLIP-GD results tend to entangle with some irrelevant attributes.
%
For example, for text prompt ``blue suit'', the generated results are entangled with ``blue eyes'', ``mouth slightly open'', ``wearing earrings'', \textit{etc}. On the contrary, DeltaEdit-G can achieve more disentangled results without complex hyper-parameters tuning.

\subsubsection{Quantitative Comparison}
\begin{table*}[t]
    \centering\scriptsize
    \caption{%
        Quantitative comparison results of GAN-based methods.
    }
    \label{tab:quatita}
    \setlength{\tabcolsep}{28pt}
    \begin{tabular}{l|ccc}
	\toprule
        Methods & FID ($\downarrow$) & PSNR ($\uparrow$) &IDS ($\uparrow$) \\
        \midrule   
        TediGAN  &$31.13$ &$20.46$ &$0.60$\\
        StyleCLIP-LM  &$18.33$ &$21.41$ &$0.88$\\
        StyleCLIP-GD ($\rho=2$) &$12.06$ &$22.31$ &$0.86$\\
        StyleCLIP-GD ($\rho=3$) &$16.85$ &$19.50$ &$0.77$\\
        StyleCLIP-GD ($\rho=4$) &$22.85$ &$17.61$ &$0.68$\\
        StyleMC &$48.96$ &$17.47$ &$0.52$\\
        \midrule
        DeltaEdit-G (Ours)&$\mathbf{10.29}$ &$\mathbf{22.92}$ &$\mathbf{0.90}$\\
        DeltaEdit-G ($\mathcal{S}$ space) &$18.88$ &$13.49$ &$0.61$\\
        DeltaEdit-G ($\mathcal{W+}$ space) &$21.63$ &$12.83$ &$0.55$\\
        DeltaEdit-G (Straightforward) &$44.62$ &$11.21$ &$0.29$\\
        \bottomrule
    \end{tabular}
\end{table*}

In Tab.~\ref{tab:quatita}, we present objective measurements of FID, PSNR, and IDS (identity similarity before and after editing by Arcface~\cite{deng2019arcface}) for GAN-based methods comparison. All results are the average on ten given texts. Compared with the state-of-the-art GAN-based methods, our DeltaEdit-G achieves the best performance on all metrics.

We also compare DeltaEdit-G with the straightforward solution to text-free training in Tab.~\ref{tab:quatita}. For fairness, the straightforward solution is also trained in the $\mathcal{S}$ space and supervised by the same loss functions as the final solution. Compared with the straightforward solution, the final solution (ours) can largely improve the inference performance on all metrics, benefiting from the well-aligned CLIP Delta image-text space.

\begin{table}[t]
    \caption{%
        User preference study on editing accuracy (Acc.) and visual realism (Real.)
    }
    \label{tab:user}
    \centering\footnotesize
    \setlength{\tabcolsep}{3pt}
    \begin{tabular}{c|cccc}
        \toprule
        & \makecell[c]{StyleCLIP \\ $\rho=2$} & \makecell[c]{StyleCLIP \\ $\rho=3$} & \makecell[c]{StyleCLIP \\ $\rho=4$} & \makecell[c]{DeltaEdit-G \\ (Ours)} \\
        \midrule   
        Acc. ($\uparrow$) &$19\%$ & $9\%$ & $9.75\%$ & $\mathbf{62.25\%}$ \\
        \midrule
        Real. ($\uparrow$) &$50\%$ & $33\%$ & $28.75\%$ & $\mathbf{90.25}$\% \\
        \bottomrule
    \end{tabular}
\end{table}

Finally, we conduct human subjective evaluations upon the editing accuracy (Acc) and visual realism (Real). We compare DeltaEdit-G with the strong GAN-based baseline, StyleCLIP-GD, under different hyper-parameter $\rho$ settings. In total, 20 evaluation rounds are performed and 40 participants are invited. At each round, we present results of randomly sampled editing text to each participant. Participants were asked to choose the best text-guided image editing output considering Acc and to select the output images (not limited to 1 image) which are visually realistic (Real). The results are listed in Tab.~\ref{tab:user}, showing the superiority of our model.

\subsection{Results of DeltaEdit-D}\label{sec:res_d}

\subsubsection{Qualitative Results}

\noindent\textbf{Semantically Meaningful Latent Interpolation.}
Our style-conditioned diffusion model encodes the image semantics into latent codes $s$. Here, we seek to explore whether simple linear changes within this latent space can effectively lead to corresponding semantic alterations in the generated images. To realize a linear interpolation effect, we encode two input images into $(s_1, x^1_T)$ and $(s_2, x^2_T)$, and subsequently apply linear interpolation to the latent code $s$ and spherical linear interpolation to the stochastic vector $x_T$, following~\cite{song2020denoising,preechakul2022diffusion}. Concretely, the linear interpolation can be mathematically represented as $s^{\lambda} = \lambda \cdot s_1 + (1-\lambda) \cdot s_2$, while the formulation of spherical linear interpolation is as follows:
\begin{equation}
    x_T^{\lambda}=\frac{\sin ((1-\lambda) \theta)}{\sin (\theta)} x_T^{1}+\frac{\sin (\lambda \theta)}{\sin (\theta)} x_T^{2},
\end{equation}
where $\theta=\arccos \left(\frac{\left(x_T^{1}\right)^{\top} x_T^{2}}{\left\|x_T^{1}\right\|\left\|x_T^{2}\right\|}\right)$, and $\lambda$ changes from 0 to 1 in intervals of 0.2.

\begin{figure}[t]
    \centering
    \includegraphics[width=1\linewidth]{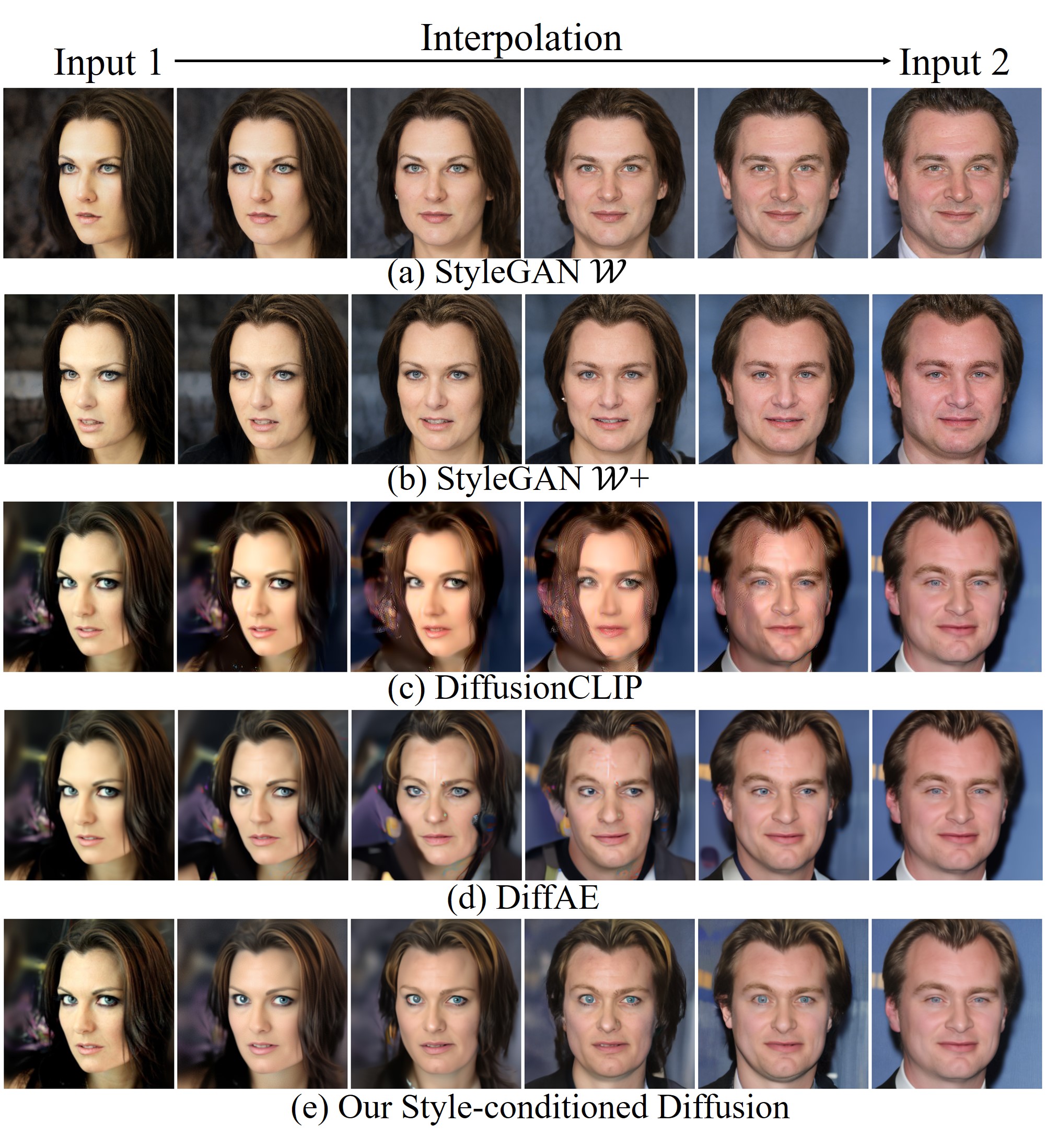}
    \caption{%
        Interpolation between two real images with different semantic spaces. Our style-conditioned diffusion approach stands out by generating interpolation results that are smooth and maintain details.
    }
    \label{fig:interpolation}
\end{figure}

\begin{table*}[t]
    \caption{%
        Quantitative results of image reconstruction using SSIM, LPIPS, and MSE metrics.
    }
    \label{tab:quatita_rec}
    \centering\scriptsize
    \setlength{\tabcolsep}{20pt}
    \begin{tabular}{l|c|ccc}
        \toprule
        Methods & Setting & SSIM ($\uparrow$) & LPIPS ($\downarrow$) &MSE ($\downarrow$) \\
        \midrule   
        StyleGAN2 ($\mathcal{W}$) &-  &$0.677$ &$0.168$ &$1.60e^{-2}$\\
        StyleGAN2 ($\mathcal{W+}$) &- &$0.827$ &$0.114$ &$6.00e^{-3}$\\
        VQ-GAN &- &$0.782$ &$0.109$ &$3.61e^{-3}$\\
        VQ-VAE2 &- &$0.947$ &$0.012$ &$4.87e^{-4}$\\
        HFGI &- &$0.877$ &$0.127$ &$6.17e^{-2}$\\
        DDIM &$T=100, 128^2$ &$0.917$ &$0.063$ &$2.00e^{-3}$\\
        DiffAE &$T=100, 128^2, x_{T}$ &$\mathbf{0.991}$ &$0.011$ &$\pmb{6.07e^{-5}}$ \\
        \midrule
        Ours &$T=100, 128^2, x_{T}$ &$0.989$ &$\mathbf{0.007}$ &$6.34e^{-5}$ \\
        \bottomrule
    \end{tabular}
\end{table*}

As shown in Fig.~\ref{fig:interpolation}, the interpolation results from different spaces of StyleGAN (the case of (a) and (b)) are visually smooth and continuous, indicating the semantic consistency within the StyleGAN latent space. Meanwhile, in the case of (b), as the StyleGAN $\mathcal{S}$ space is solely obtained through affine transformations of the $\mathcal{W+}$ space, both exhibit the same interpolation effect. However, StyleGAN's weak ability to reconstruct local details, such as background and facial attributes, results in interpolation results that do not accurately match the real input image. 
Differently, methods based on the diffusion model, including (c), (d), and (e), can offer high-quality reconstructions of real images through deterministic versions, thus enabling more accurate matching of the real input images. Among them, DiffusionCLIP directly edits the random noise latent space to generate a series of interpolation results. Yet, noise latent variables cannot accurately capture high-dimensional semantics, leading to degraded outcomes. Although DiffAE achieves smoother interpolation results compared to DiffusionCLIP, its encoded semantic latent vectors merely represent a holistic vector, possibly resulting in insufficient semantic representation and causing some interpolation effects to remain non-smooth. 
In contrast, our diffusion approach directly employs the semantically rich StyleGAN latent space to control the diffusion model with detailed reconstruction capabilities. This allows for gradual changes in head pose, background, and facial attributes between the given images without introducing artifacts.

\noindent\textbf{Real Image Reconstruction.}
The quality of image reconstruction can reflect the accuracy of a generative model in encoding-decoding real images. As shown in Tab.~\ref{tab:quatita_rec}, we evaluate image reconstruction accuracy of various methods, including: (1) GAN inversion methods (including pretrained StyleGAN with $\mathcal{W}$ or $\mathcal{W+}$ sapce~\cite{karras2020analyzing} as well as VQ-GAN~\cite{esser2021taming}); (2) VAE-based methods (such as VQ-VAE2~\cite{razavi2019generating}); (3) diffusion-based methods (including DDIM~\cite{song2020denoising}, DiffAE~\cite{preechakul2022diffusion}, and our proposed diffusion model). All models are trained on the FFHQ~\cite{karras2019style} dataset and tested on the CelebA-HQ~\cite{karras2017progressive} 30K dataset. All diffusion-based methods use T=100 for decoding. Our diffusion method and the diffusion-based DiffAE approach demonstrate superior performance, surpassing previous GAN-based and VAE-based methods by a large margin, thus confirming their near-perfect reconstruction capabilities.

\begin{figure*}[t]
    \centering
    \includegraphics[width=1\linewidth]{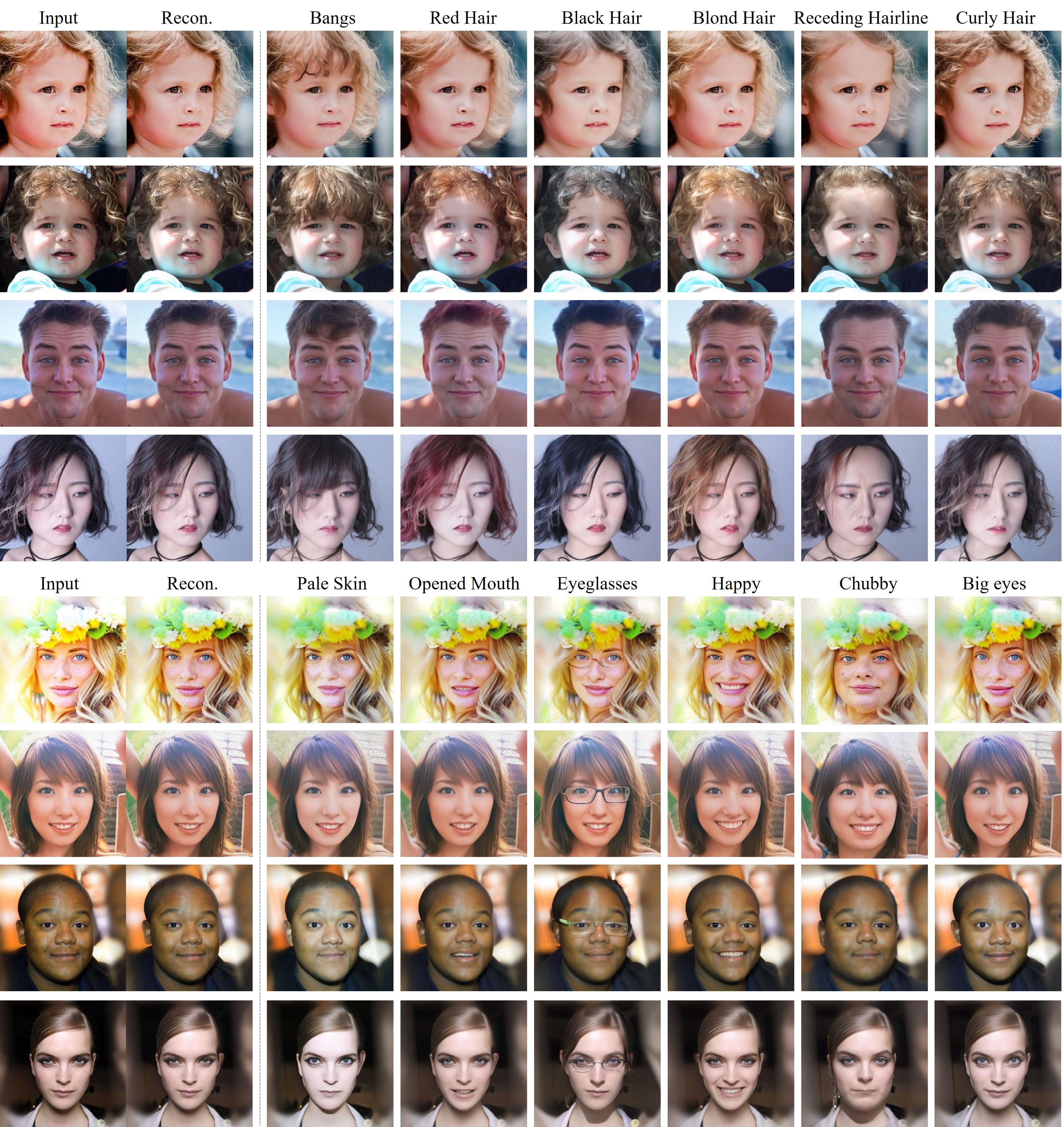}
    \caption{%
        Results of DeltaEdit-D for facial images, using our style-conditioned diffusion model trained on the FFHQ dataset. The target attribute included in the text prompt is above each image.
    }
    \label{fig:face_diffusion}
\end{figure*}

\noindent\textbf{Text-guided Image Editing.}
Unlike editing with StyleGAN, details often change due to the inability to accurately invert real images to the latent space of it. The advantage of editing with the diffusion model lies in its ability to edit real images while preserving unrelated details unchanged. As shown in Fig.~\ref{fig:face_diffusion}, we demonstrate reconstructions of real face images and 12 editing results guided by different text prompts. Our DeltaEdit-D based on the proposed Style-conditioned diffusion can almost perfectly reconstruct real images. Additionally, given texts with various semantic meanings, our approach achieves accurate image editing effects while maintaining the identity, background, and other unrelated attributes of the facial images. Note that our proposed diffusion model is trained only on FFHQ~\cite{karras2019style}, yet we test its editing performance on both FFHQ (upper subfigure of Fig.~\ref{fig:face_diffusion}) and CelebA-HQ~\cite{karras2017progressive} (lower subfigure of Fig.~\ref{fig:face_diffusion}). This indicates that our method generalizes well to CelebA-HQ without the need for fine-tuning.



\begin{figure*}[t]
    \centering
    \includegraphics[width=1\linewidth]{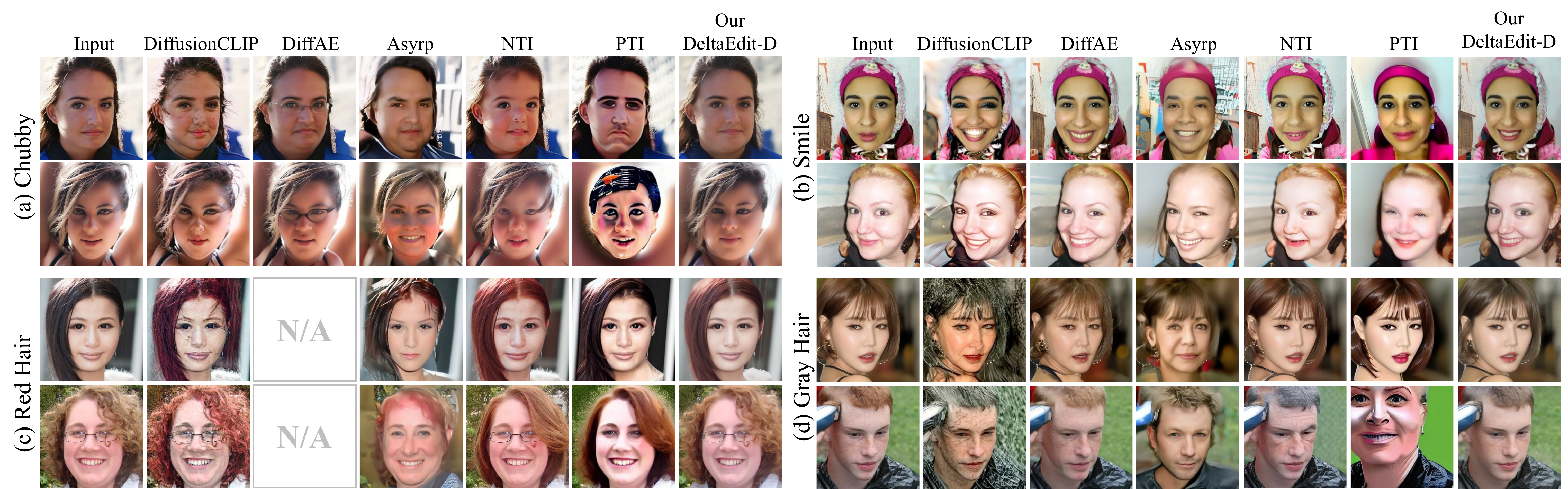}
    \caption{%
        Comparison results with diffusion-based methods, including DiffusionCLIP, DiffAE Asyrp, NTI and PTI. Our DeltaEdit-D approach shows better visual realism and attribute disentanglement almost in all cases.
    }
    \label{fig:com_diffusion}
\end{figure*}

\subsubsection{Qualitative Comparison}

We conduct a comparison with diffusion-based image editing methods, categorizing them into two distinct approaches: (1) Category-specific methods trained from scratch, including DiffusionCLIP~\cite{kim2022diffusionclip}, DiffAE~\cite{preechakul2022diffusion}, and Asyrp~\cite{kwon2022diffusion}, which require retraining the model for each target category. (2) Textual inversion-based editing methods using pre-trained Stable Diffusion~\cite{rombach2022high}, including NTI~\cite{mokady2023null} and PTI~\cite{dong2023prompt}, which leverage the generalization capability of pre-trained models through latent space inversion.

\begin{table}[t]
    \caption{%
        Quantitative comparison results of DeltaEdit-D with diffusion-based methods.
    }
    \label{tab:quatita_diffusion}
    \centering\footnotesize
    \small
    \setlength{\tabcolsep}{4pt}
    \begin{tabular}{l|ccc}
	\toprule
        Methods & FID ($\downarrow$) & PSNR ($\uparrow$) &IDS ($\uparrow$) \\
        \midrule
        DiffusionCLIP  &72.21 &22.08 &0.73\\
        DiffAE  &15.73 &23.40 &0.84\\
        Asyrp &87.38 &20.20 &0.48\\
        \midrule
        DeltaEdit-D (Ours) &$\mathbf{9.36}$ &$\mathbf{23.51}$ &$\mathbf{0.93}$\\
        \bottomrule
    \end{tabular}
\end{table}

\begin{figure}[t]
    \centering
    \includegraphics[width=1\linewidth]{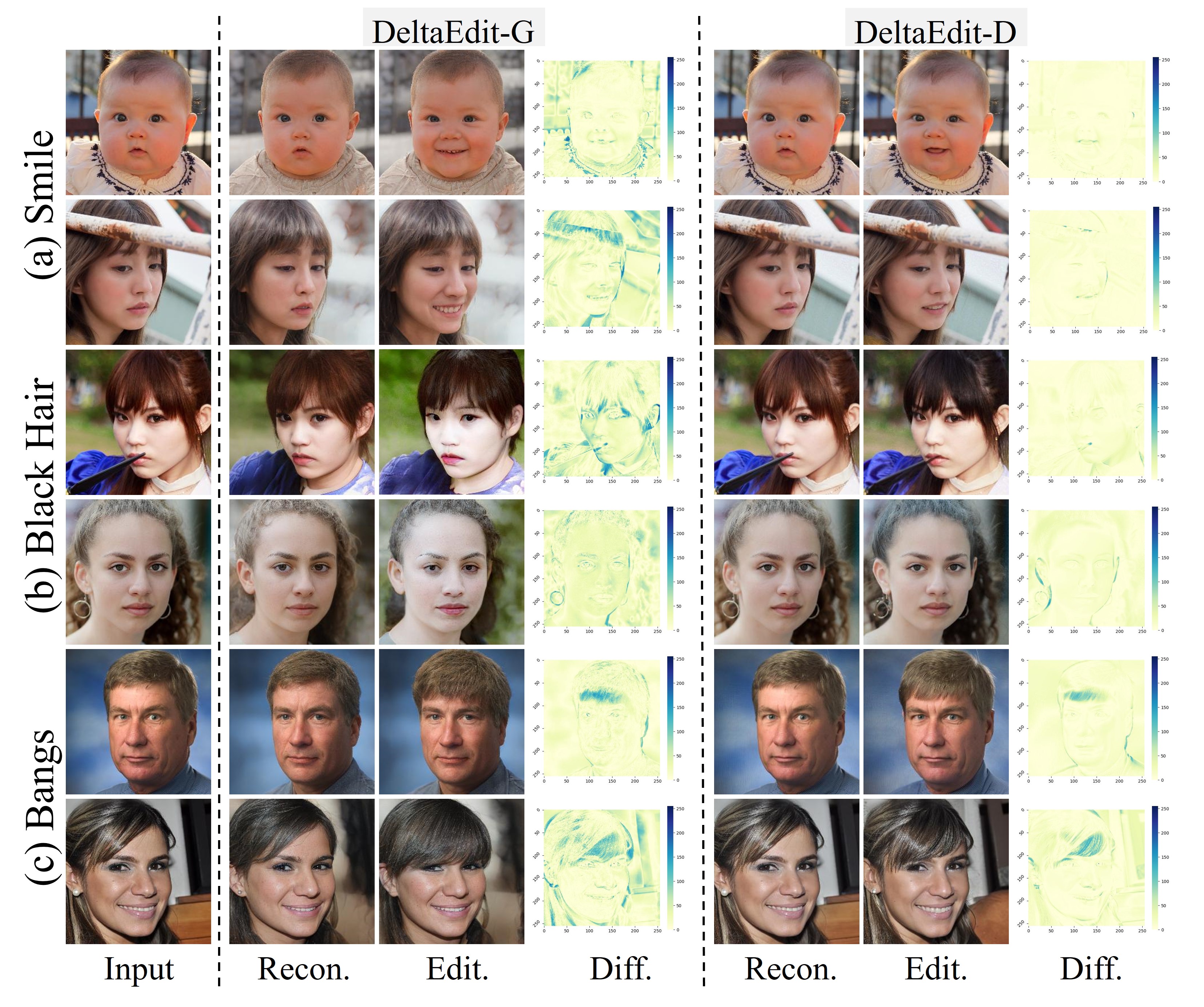}
    \caption{Comparison results between DeltaEdit-G and DeltaEdit-D. ``Recon.'' stands for the results of image reconstruction from the input image. ``Edit.'' represents the editing results corresponding to the provided texts. ``Diff.'' indicates the heatmaps of the $\mathcal{L}_1$ differences between ``Edit.'' and ``Input'', demonstrating that editing with DeltaEdit-D has minimal impact on unrelated regions.}
    \label{fig:two_ours}
\end{figure}

\begin{figure*}[t]
    \centering
    \includegraphics[width=1\linewidth]{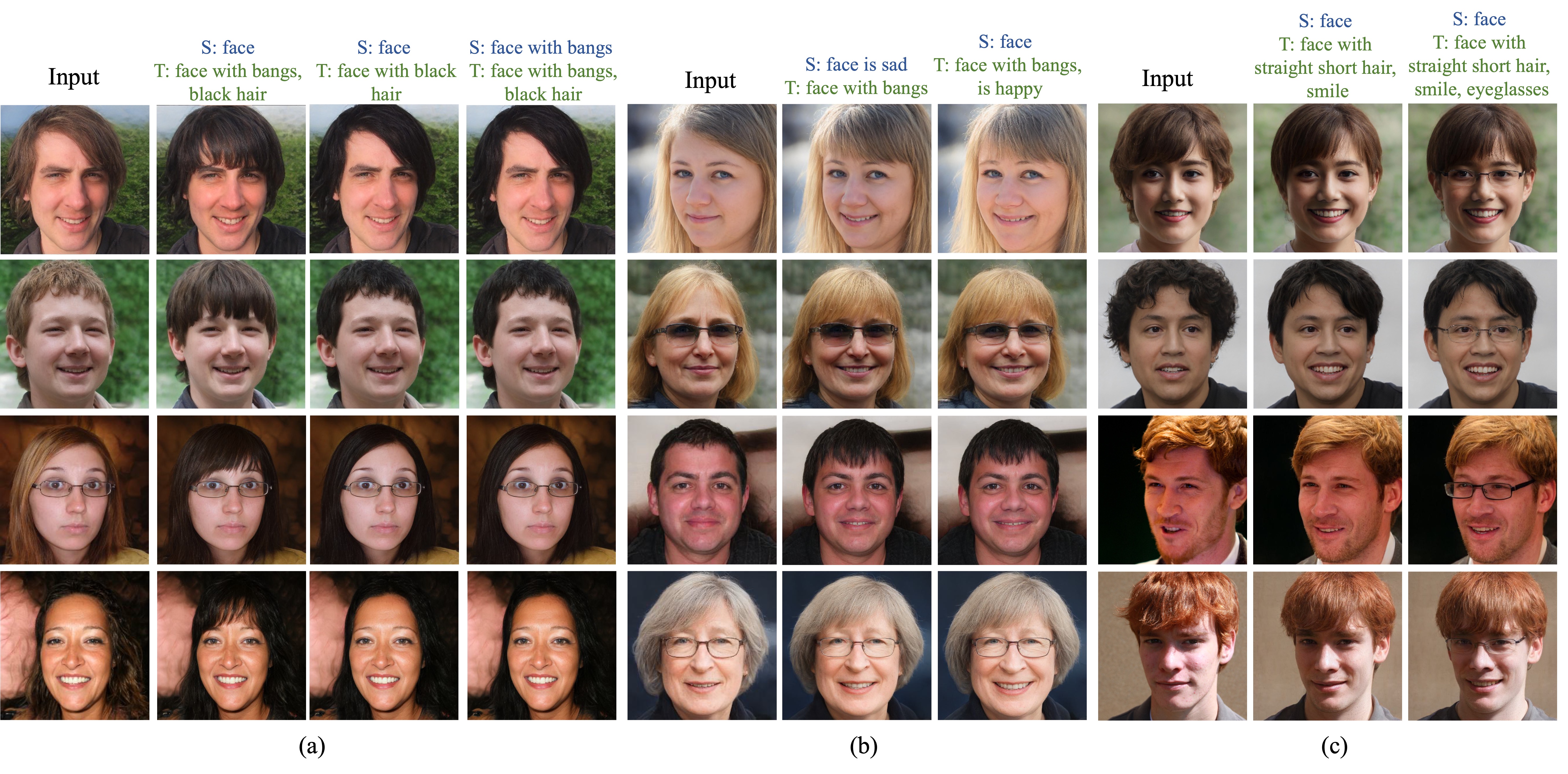}
    \caption{%
        Editing results under different text prompt settings, where the source text is labeled blue and the target text is labeled green. \newtext{Among these, the first two rows show the results from DeltaEdit-G, while the last two rows display the results from DeltaEdit-D. Testing across different models can reveal the effectiveness of the method under varying text settings.}
    }
    \label{fig:prompt}
\end{figure*}

As shown in Fig.~\ref{fig:com_diffusion}, category-specific methods such as DiffusionCLIP and Asyrp display corresponding modifications to target attributes provided by texts. However, these results exhibit low quality and have entanglement issues. For example, when targeting ``gray hair'', both DiffusionCLIP and Asyrp incorrectly modify the identity and background of the edited faces. The results of DiffAE exhibit higher editing quality than DiffusionCLIP and Asyrp. Nonetheless, DiffAE still suffers from entanglement issues. For example, when editing for ``chubby'', the results may wrongly incorporate the attribute of wearing glasses. Editing for ``gray hair'' results in hair color alteration alongside face aging. Furthermore, DiffAE performs editing by shifting the latent vector linearly along the target direction, which is found by training a linear classifier (logistic regression) to predict the target attribute using a labeled dataset. This process leads to limitations, preventing DiffAE from achieving the corresponding editing for unannotated attributes such as ``red hair'', as shown in Fig.~\ref{fig:com_diffusion} (c). In contrast, our approach does not rely on images' attribute annotations, enabling flexible attribute editing based on any given text while maintaining high quality and disentanglement capability.

Among textual inversion-based editing methods, NTI~\cite{mokady2023null} effectively preserves background details in input images through its noise-space null-text inversion. When combined with P2P~\cite{hertz2023prompt}'s attention-level localization, it achieves precise control over specific editing regions in most samples. In contrast, PTI~\cite{dong2023prompt} encodes input images into the text space, which inevitably causes content loss from the original input and consequently significantly compromises editing outcomes. Our method demonstrates superior performance by achieving high-quality editing results across all samples. It employs DDIM inversion to preserve details while utilizing style-conditioned encoding for semantic control. Unlike textual inversion methods that require per-sample optimization, our approach delivers stable, high-quality editing after a single training phase.

\subsubsection{Quantitative Comparison}

In Tab.~\ref{tab:quatita_diffusion}, we present objective measurements of FID, PSNR, and IDS (identity similarity before and after editing by Arcface~\cite{deng2019arcface}) for diffusion-based methods comparison. 
The results are the average on ten given texts.
The quantitative results have the best  performance across multiple metrics. These results are consistent with the qualitative results that our DeltaEdit-D method exhibits the best disentanglement and identity preservation compared to diffusion-based models.

\subsection{Comparison between DeltaEdit-G and DeltaEdit-D}\label{sec:com_d}
\begin{table*}[t]
    \caption{%
        Time efficiency comparison of our method and other state-of-the-art methods. $*$ means that the additional time for manually hyper-parameters tuning during the inference is excluded, which is typically 8-9 seconds for each case.
    }
    \label{table:time}
    \scriptsize
    \setlength{\tabcolsep}{12pt}
    \begin{tabular}{l|c|c|c|c|c}
        \toprule
        & \makecell[c]{Pre-processing \\ time} & \makecell[c]{Training \\ time} & \makecell[c]{Inference \\ time} & \makecell[c]{Conditioned on \\ input image} & \makecell[c]{Latent \\ space} \\
        \midrule   
        TediGAN &- & 12h+ & 20 s & yes & $\mathcal{W+}$ \\
        StyleCLIP-OP &- & - & 99 s & yes & $\mathcal{W+}$ \\
        StyleCLIP-LM &- & 10-12h & 70ms & yes & $\mathcal{W+}$ \\
        StyleCLIP-GD &4h & - & 72ms$^*$ & no & $\mathcal{S}$ \\
        StyleMC &- & 5s & 65ms & yes & $\mathcal{S}$ \\
        \midrule
        DeltaEdit-G (Ours) &4h & 2.7h & 73ms & yes & $\mathcal{S}$ \\
	\newtext{DeltaEdit-D (Ours)} &\newtext{-} & \newtext{12h+} & \newtext{14.6s} & \newtext{yes} & \newtext{$\mathcal{S}$} \\
	\bottomrule
    \end{tabular}
\end{table*}


Our identified DeltaSpace and proposed DeltaEdit can be applied across different generative models such as StyleGAN and the diffusion model, allowing for controllable and flexible editing. In this section, we compare the results obtained by instantiating DeltaEdit on two types of generative models (DeltaEdit-G and DeltaEdit-D). As depicted in Fig.~\ref{fig:two_ours}, we reconstruct multiple input images using StyleGAN and the diffusion model as decoders. Notably, reconstructions based on the diffusion model exhibit superior accuracy compared to those based on StyleGAN, attributed to the diffusion model's advantage in encoding details in the noise space. Additionally, we utilize the trained DeltaEdit model to generate corresponding latent space editing directions for different input texts. These directions are then applied to StyleGAN (DeltaEdit-G approach) and the proposed diffusion model (DeltaEdit-D approach) to generate edited images, denoted as ``Edit.''. Moreover, to provide a clear comparison of the editing effects between DeltaEdit-G and DeltaEdit-D, we compute and display the heatmaps of $\mathcal{L}_1$ differences between the edited results (``Edit.'') and input images, labeled as ``Diff.''. The pixel values of the images are all within the range of [0-255] when calculating the $\mathcal{L}_1$ differences. By comparing the $\mathcal{L}_1$ differences heatmaps of DeltaEdit-G and DeltaEdit-D, we find that both methods are capable of attribute editing and affecting relevant regions. However, due to the inability of StyleGAN's inversion network to accurately invert real images to latent space, modifications along the target editing direction for this latent code can change unrelated areas in the generated images. This further demonstrates that our DeltaEdit can serve as an effective tool for analyzing and understanding the generative and editing capabilities of generative models.

\begin{figure}[t]
    \centering
    \includegraphics[width=1\linewidth]{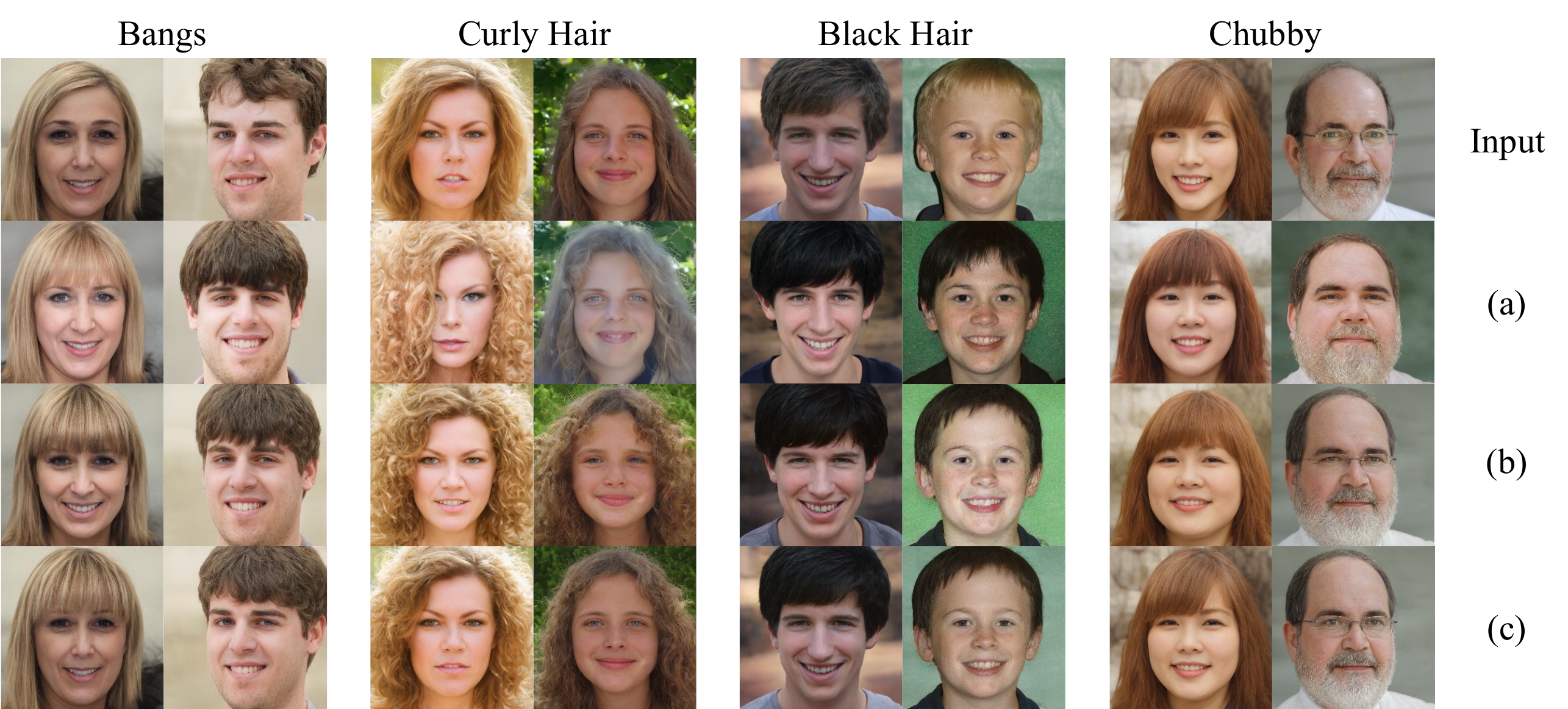}
    \caption{%
        Editing results by implementing our DeltaEdit in (a) W+ space, (b) S space, and (c) S space with relevance matrix $R_s$. \newtext{For each attribute, the first column shows DeltaEdit-G results, and the second column displays DeltaEdit-D results.}
    }
    \label{fig:ablation}
\end{figure}



\subsection{Model Analysis}

\subsubsection{Text-prompt Setting}

In Fig.~\ref{fig:prompt}, we explore how different text-prompt settings can affect the editing results \newtext{by presenting the outputs of both DeltaEdit-G and DeltaEdit-D}. During inference, the editing direction $\Delta \hat s$ is driven by the CLIP text space direction $\Delta t$ between source text and target text.
(a) We first construct three different (source text, target text) pairs and find that regardless of the content of them, the editing results are affected by the difference between them. For example, although both the source and target texts contain the attribute of ``bangs'', the editing result is only influenced by the difference, namely ``black hair''.
(b) Moreover, to align with the training phase, we construct (source text, target text) pairs using the specific text description for the image, such as “face is sad” as the source text and “face with bangs” as the target text. We find that the generated faces become more ``happy'' (the opposite direction of the source text) and with ``bangs'' (the direction of the target text), which is equivalent to directly putting the attributes to be edited all into the target text.
(c) In addition, we construct target texts with multiple combinations of facial attributes, including hairstyles, smile and eyeglasses, and our method can yield desired results driven by text prompts containing multiple semantics. Note that our method can directly perform multi-attributes editing without additional training processes, since the different editing directions have been learned well by training on large-scale data.

\subsubsection{Efficiency Analysis}

To validate the efficiency and flexibility of the proposed method, we compare the computation time with TediGAN, StyleMC and three StyleCLIP approaches in Tab.~\ref{table:time}.
Specifically, TediGAN first requires 12+ hours to encode images and texts into a common space and then trains the encoding module.
StyleCLIP-OP manipulates images with an optimization process, which requires several tens of seconds. 
StyleCLIP-LM is fast during inference, but needs 10-12 hours to train a mapper network for per text prompt. 
%
StyleCLIP-GD requires about 4 hours to pre-compute the global edit directions. However, since the pre-computed directions are rough and cannot be directly applied to control the manipulation, it requires additional manual tuning for different text prompts, which typically takes 8-9 seconds for each case. 
Differently, StyleMC is fast during training. However, for N text prompts, StyleMC still needs N times longer to train the corresponding model. 
On the contrary, our \newtext{DeltaEdit-G} method avoids the labor time during training and inference. Once trained, our Delta Mapper is universal and can directly manipulate images with new text prompts efficiently. Meanwhile, our Delta Mapper only needs 2-3 hours for training, since the training is directly conducted on the latent space without generating images in each iteration. \newtext{However, it is worth noting that DeltaEdit-D requires additional training time to learn the style-based diffusion model, resulting in a longer training phase compared to DeltaEdit-G. Additionally, since DeltaEdit-D relies on iterative denoising via the diffusion model to generate edited results, its inference time is also longer. Nevertheless, this progressive denoising process enables DeltaEdit-D to achieve higher-quality image edits than DeltaEdit-G, as demonstrated in Fig.~\ref{fig:two_ours}.}

\subsubsection{Choice of the Editing Space}
To find an appropriate space for the proposed method, we conduct experiments by performing editing in $\mathcal{W+}$ space and $\mathcal{S}$ space, \newtext{validating the results on both GAN and diffusion models.}. 
The quantitative and qualitative results in Tab.~\ref{tab:quatita} and Fig.~\ref{fig:ablation} show that $\mathcal{S}$ results can achieve better visual quality and identity preservation than $\mathcal{W+}$.
For instance, in Fig.~\ref{fig:ablation}, when considering the attribute ``bangs'', $\mathcal{W+}$ results succeed in adding bangs to given faces. However, the attribute-irrelevant contents are heavily changed, such as poses, skin color, and background. In contrast, $\mathcal{S}$ results have more disentanglement than $\mathcal{W+}$, \emph{e.g.}, the poses, facial identities are well preserved. 
%
%
Thus, we implement our DeltaEdit in $\mathcal{S}$ space, and introduce $R_s$ in the disentanglement part of Sec.~\ref{sec:solution2} to further improve the disentanglement performance.

\subsubsection{Effectiveness of Relevance Matrix $R_s$}
For further improving the disentanglement, we introduce relevance matrix $R_s$ to limit some irrelevant channels from changing. The fourth row in Fig.~\ref{fig:ablation} shows that, with $R_s$, our method can successfully edit the desired attributes while preserving the text-irrelevant content unchanged. For example, for ``bangs'', the generated results have added bangs while accurately preserving the background and facial pose. 

\section{Conclusion}\label{sec:conclusion}

In this paper, we have introduced DeltaEdit, a flexible framework for text-guided image editing. The core of our approach has been the discovery and utilization of the well-aligned CLIP DeltaSpace. During training, the framework has learned to map differences in CLIP visual features to corresponding directions in a generative model's latent space. For inference, it then leverages differences in CLIP text features to predict these latent directions. This design has enabled training without reliance on costly image-text pair collections and ensures strong generalization to unseen text prompts for zero-shot editing. Extensive qualitative and quantitative experiments across different generative models, including both GANs and diffusion models, have demonstrated the superiority of DeltaEdit. Our method has shown compelling advantages in generating high-quality results, achieving efficiency in both training and inference, and maintaining robust generalization to arbitrary, unseen textual descriptions.


%


\bibliography{sn-bibliography}

\end{document}